\documentclass[letterpaper]{article} 
\usepackage{aaai2026}  
\usepackage{times}  
\usepackage{helvet}  
\usepackage{courier}  
\usepackage[hyphens]{url}  
\usepackage{graphicx} 
\usepackage{natbib}  
\usepackage{caption} 
\usepackage{algorithm}
\usepackage{algorithmic}

\usepackage{newfloat}
\usepackage{listings}
\DeclareCaptionStyle{ruled}{labelfont=normalfont,labelsep=colon,strut=off} 
\floatstyle{ruled}
\newfloat{listing}{tb}{lst}{}
\floatname{listing}{Listing}
\title{Urban Incident Prediction with Graph Neural Networks: Integrating Government Ratings and Crowdsourced Reports}
\author{
    Sidhika Balachandar\textsuperscript{\rm 1}\thanks{Correspondence to sidhikab@cs.cornell.edu or\\ ngarg@cornell.edu},
    Shuvom Sadhuka\textsuperscript{\rm 2},
    Bonnie Berger\textsuperscript{\rm 2},
    Emma Pierson\textsuperscript{\rm 1},
    Nikhil Garg\textsuperscript{\rm 3}
}
\affiliations{
    \textsuperscript{\rm 1} Department of Electrical Engineering \& Computer Sciences, UC Berkeley\\
    \textsuperscript{\rm 2} Department of Electrical Engineering \& Computer Science, MIT\\
    \textsuperscript{\rm 3} Department of Operations Research \& Information Engineering, Cornell Tech
}

\usepackage{tabularx}
\usepackage{amssymb}
\usepackage{amsmath}
\usepackage{booktabs}
\usepackage{multirow}
\usepackage{amsfonts}
\usepackage{dsfont}

\newif\ifcomments
 \commentstrue
\ifcomments
\newcommand\emma[1]{\textcolor{blue}{\textbf{EP}: #1}}
\newcommand\nikhil[1]{\textcolor{teal}{\textbf{NG}: #1}}

\else
    \newcommand\emma[1]{}
    \newcommand\nikhil[1]{}
\fi
\newcommand\node{i}
\newcommand\Tikt{\ensuremath{T_{\node kt}}}

\newcommand\rikt{\ensuremath{r_{\node kt}}}

\newcommand\bbR{\ensuremath{\mathbb{R}}}
\newcommand\bbE{\ensuremath{\mathbb{E}}}

\begin{document}

\maketitle

\begin{abstract}
Graph neural networks (GNNs) are widely used in urban spatiotemporal forecasting, e.g., predicting infrastructure problems. In this setting, government officials aim to identify in which neighborhoods incidents like potholes or rodents occur. The true state of incidents is observed via government inspection \textit{ratings}. However, these ratings are only conducted for a sparse set of neighborhoods and incident types. We also observe the state of incidents via crowdsourced \textit{reports}, which are more densely observed but may be biased due to heterogeneous reporting. First, we propose a multi-view, multi-output GNN-based model that uses both unbiased rating data and biased reporting data to predict the true latent state of incidents. Second, we investigate a case study of New York City urban incidents and collect a dataset of 9,615,863 crowdsourced reports and 1,041,415 government inspection ratings over 3 years and across 139 types of incidents. We show on both real and semi-synthetic data that our model can better predict the latent state compared to models that use only reporting data or only rating data. Finally, we quantify demographic biases in crowdsourced reporting, e.g., higher-income neighborhoods report problems at higher rates. Our analysis showcases a widely applicable approach for latent state prediction using heterogeneous, sparse, and biased data.  
\end{abstract}

\begin{links}
    \link{Model}{https://github.com/sidhikabalachandar/nyc_urban_incident_model}
    \link{Dataset}{https://github.com/sidhikabalachandar/nyc_urban_incident_data}
\end{links}

\section{Introduction}
\label{sec:introduction}
Graph neural networks (GNNs) have emerged as expressive models for making predictions on graph-structured data, including for urban applications such as air quality monitoring, forecasting traffic flows, etc. \citep{xie2019sequential,roy2021sst,brimos2023explainable, yu2023spatio, zhan2024neural}. In urban planning -- our empirical setting -- government officials often wish to know where incidents like rodent infestations or floods occur so they can make downstream resource allocation decisions; however, ground truth is often unobserved. GNNs are a powerful prediction tool, as they can encode spatial correlations of the ground-truth states across nodes in a graph (e.g., neighborhoods in a city) as well as across hundreds of types of incidents. For example, if a flood has occurred in one neighborhood, the adjacent neighborhoods are also likely to be flooded and might also experience related incidents like street damage. 

Estimating latent ground truth for the hundreds of types of incidents in a city is challenging. Two sources of information are often used, each with its limitations. First, we observe the ground-truth state via \textit{government inspections}, which generate \textit{ratings} for neighborhoods. Importantly, these inspections are only conducted for some incident types and neighborhoods and thus ratings are sparsely observed. For example, New York City conducts street inspections for every street and rates them from 1-10, but each street is only rated about once every year. Second, we observe biased proxies of the latent state via crowdsourced \textit{reports} of incidents. Compared to ratings, indicators of whether a report is observed are available for every incident type and neighborhood.

A challenge is that learning only from reporting data leads to biased predictions. Previous work has established that underreporting is pervasive and heterogeneous \citep{o2015ecometrics,o2017uncharted,clark2020advanced,kontokosta2021bias, agostini2024bayesian, liu2024quantifying}; in different neighborhoods that face similar incidents, residents often \textit{report} those incidents at different rates. This bias presents an identifiability issue; if one neighborhood logs more reports than another, it is unclear whether the former has a worse ground truth, or if given the same ground truth, the latter is less likely to report. As a result, for example, \citet{casey2018cautionary} found that in Washington, D.C., crowdsourced reports on rodents did not accurately predict the outcome of inspections. Moreover, differences in reporting often correlate with demographics, so learning only from reporting data risks introducing decision making disparities against underserved populations \cite{agostini2024bayesian, liu2024quantifying}. 

\begin{figure}
\vspace{-0em}
  \begin{center}
    \includegraphics[width=0.47\textwidth]{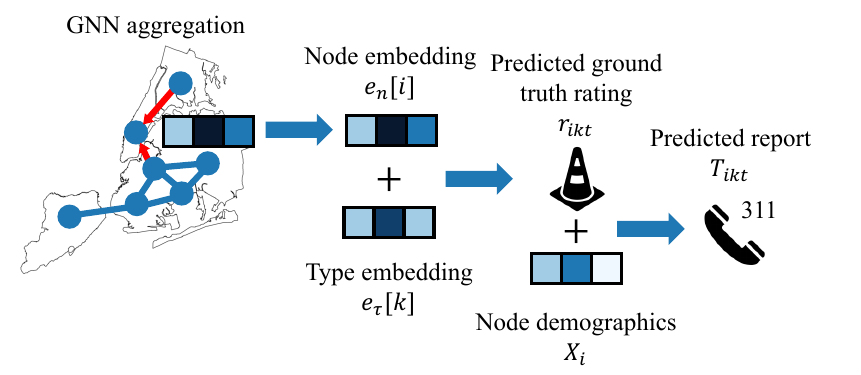}
  \end{center}
  \caption{\textit{Model.} We introduce URBAN (Unbiased Ratings and Biased reports Aggregation Network), a GNN-based model to estimate inspection ratings and reports of incidents. We model inspection ratings $\rikt$ using node $\node$'s learned embedding $e_n[i]$ and type $k$'s learned embedding $e_\tau[k]$. We model reports $\Tikt$ as a function of the rating $\rikt$ and demographics $X_\node$.}
  \label{fig:model}
  \vspace{-1.5em}
\end{figure}

In summary, we make the following contributions:
\begin{itemize}
    \item We propose a multi-view, multi-output GNN-based model for our task named URBAN (Unbiased Ratings and Biased reports Aggregation Network). As summarized in Figure \ref{fig:model}, we use both sparsely observed ground-truth rating data and frequently observed biased reporting data to predict ratings and to infer how the likelihood of reporting varies by demographics, conditional on ground-truth. URBAN's contribution lies in adapting GNNs for biased data settings by connecting multi-view datasets through a multi-output loss. Our model is available here: \url{https://github.com/sidhikabalachandar/nyc_urban_incident_model}.
    \item We investigate a case study of New York City incidents. We create a dataset of reports sourced from NYC 311-complaints (crowdsourced reports) and ratings sourced from government inspections. Our dataset presents two advantages. First, our dataset is heterogeneous, and through various train/test splits our data can be used to evaluate model performance on unseen time periods, neighborhoods, and incident types. Second, new data for a large set of types and locations is available daily, so our data provides an ideal setting to study distribution shifts over time with an automatic source of uncontaminated test sets. We share our data here: \url{https://github.com/sidhikabalachandar/nyc_urban_incident_data}.
    \item We show that our approach can predict ground-truth inspection ratings and quantify reporting biases. In both semi-synthetic and real data, we find that jointly using the sparsely observed ground-truth rating data and the frequently observed biased reporting data outperforms using reporting data alone (predicted ratings are $5.3\times$ more correlated) or using rating data alone (predicted ratings are up to $2.6\times$ more correlated). Finally, we quantify the biases in the reporting data, finding for example that conditional on ground-truth, neighborhoods with higher median income report incidents at higher rates. 
\end{itemize}

Although our primary application is urban crowdsourcing, our approach is applicable to other tasks where both sparsely observed ground-truth data and frequently observed biased data are available. This includes other urban challenges (e.g., estimating air quality using resident reports and sparse sensor measurements \citep{sarto2016bayesian, samal2022multioutput}) and spatiotemporal processes (e.g., epidemic forecasting using internet search data and sparse official health reports \citep{kapoor2020examining, hacker2022spatiotemporal, tomy2022estimating, wang2022causalgnn, yu2023spatio}). 

\section{Related work}
\label{sec:related_work}
Our work is related to literature in three areas. We provide below a brief exposition of these areas, and provide more details in the appendix related works section.

\paragraph{Predicting urban incidents:} There is a large literature on predicting urban incidents \citep{budde2014leveraging, o2015ecometrics, lum2016predict, sarto2016bayesian, wang2017structure, hu2018urban, mosavi2018flood, yu2018spatiotemporal, gomez2019spatiotemporal, pan2019matrix, xie2019sequential, clark2020advanced, jimenez2020rapid, kontokosta2021bias, agonafir2022understanding, hacker2022spatiotemporal, mauerman2022high, farahmand2023spatial, zaouche2023bayesian, agostini2024bayesian, gao2024improve}, and prior work has shown that crowdsourced reporting data may not represent ground-truth states due to underreporting \citep{wang2017structure, clark2020advanced, kontokosta2021bias, agostini2024bayesian, gao2024improve, liu2024quantifying, liu2024redesigning}. Especially notable is the work of \citet{casey2018cautionary}, who show in Washington, D.C., that 311-reports alone are poor predictors of ground-truth ratings.

Existing works in this area typically consider only one or a few related incident types; thus a novel contribution of our model is its ability to learn and make predictions for more than a hundred 311-complaint types. Additionally, many related works use statistical models that make strict assumptions \citep{agostini2024bayesian}, whereas our GNN-based model presents a flexible approach to modeling spatial correlation. Finally, we create a heterogeneous spatial dataset that we hope can be used by other researchers to study data bias in an urban application.

\paragraph{Using GNNs for biased data:} A common approach to addressing selection bias in GNNs is \textit{graph-based semi-supervised learning} \cite{song2023graph}, which typically uses graph regularization \citep{zhou2003learning, zhu2003semi, belkin2006manifold, dai2021nrgnn, verma2021graphmix, li2023informative, qian2023robust, zhou2024varianceenlarged} or graph embedding methods \citep{perozzi2014deepwalk, tang2015line, grover2016node2vec, li2021comatch, sharma2023efficiently} to infer labels for unlabeled datapoints.

These related works discuss how to effectively train GNNs on sparsely labeled data (e.g., inspection ratings) by generating proxy labels for the unlabeled datapoints. We extend this literature by identifying another approach to obtaining proxy labels: using data from a frequently observed biased proxy for the target of interest (e.g., whether a report is made is a biased proxy of the true rating).

\paragraph{Using GNNs for spatiotemporal modeling:} GNNs are a natural fit for high-dimensional spatiotemporal modeling \citep{kapoor2020examining, roy2021sst, wang2022causalgnn, wang2022traffic, he2023stgc, pineda2023geometric, wu2024equivariant}. Our study relates to literature on \textit{multi-view GNNs}, which integrate information from multiple datasets or views \citep{shao2023heterogenous, cheng2018spatiotemporal}, and \textit{multi-output GNNs}, which predict multiple outputs from a single graph by leveraging shared graph structure and correlations between the outputs \citep{peng2022multi, samal2022multioutput}.

Related works that use GNNs for spatiotemporal modeling have identified that learning only from proxy data (reports) results in biased predictions. We extend this literature by identifying an approach to improve predictions: additionally learning from sparse ground-truth data (ratings). We show that our model, which uses both reporting and rating data, makes better predictions than a model trained only on reporting data or only on rating data. Additionally, while multi-view and multi-output GNNs are used for spatiotemporal tasks, prior work does not explicitly model biases. 
\section{Setting and data}
\label{sec:data}
We now describe our setting. The goal is to estimate the ground-truth state of various types of incidents (e.g., rodents, floods, etc.). There are two sources of data for this task: frequently observed biased crowdsourced \textit{reporting} data and sparsely observed unbiased inspection \textit{rating} data. To our knowledge, prior urban prediction work has not considered the challenges and benefits of jointly learning from these two data sources---one reason is the lack of processed data combining reports and ratings across types. We formalize this setting and provide large-scale preprocessed data composed of 9,615,863 crowdsourced reports across 139 incident types and 1,041,415 government inspection ratings across 5 types. Additional details are provided in the appendix data section.

\paragraph{Dataset overview:} We create a multi-view dataset of ratings and reports from New York City. We analyze all Census tracts\footnote{A Census tract is a geographic region defined by the U.S. Census Bureau. On average, each Census tract has thousands of inhabitants. There are $2326$ total Census tracts in NYC.} with valid demographic information ($2292$ tracts). We use all complaint types that receive at least one report for greater than 0.1\% of all tract/week pairs. We study all data in the three year period from 2021-2023 discretized into $157$ weeks. The main contribution is the data's heterogeneity. Our data can be used to evaluate models on unseen time periods, tracts, and types. Many related works only analyze \textit{one or a few} types or do not model relationships between types \cite{agostini2024bayesian,liu2024quantifying, liu2024redesigning, xu2017predicting, clark2020advanced, kontokosta2017equity, agonafir2022understanding, casey2018cautionary}. In contrast, our data represents the majority of available types and our model can learn relationships between types. 

\paragraph{Reports:} We observe reports of incidents from New York City's resident reporting system, NYC311 \citep{311data}. We collect data from 9,615,863 reports. We transform reports into \textit{indicators} of whether a report of a type was observed in a Census tract during a particular week. 

\paragraph{Ratings:} For some Census tract, type, and week tuples, we observe inspection ratings. We source ratings from government inspections (e.g., street ratings, park ratings, etc.). A lower rating indicates a worse state; e.g., a street with a lower rating has more damage. We normalize ($z$-score) the ratings for each type across time and Census tracts. 

We collect ratings from 1,041,415 inspections across five types: (i) street conditions \citep{street_ratings}, (ii) park maintenance or facility conditions \citep{park_ratings}, (iii) rodents \citep{rodent_ratings}, (iv) food establishment/mobile food vendor/food poisoning \citep{restaurant_ratings}, and (v) DCWP consumer complaints \citep{dcwp_ratings}. We \textit{do not} observe ratings for every tract/week pair. A summary is in the appendix data section. 

Our model treats government inspections as a source of unbiased data, so we ensure that the inspections are not susceptible to bias. Selection bias can occur if an inspection is conducted in response to a crowdsourced report. We process the inspection data to only use inspections \textit{not} conducted in response to crowdsourced reports, as verified by the data dictionaries or by filtering out responsive inspections. For example, rodent inspections are conducted both block-by-block in a scheduled manner and in response to 311-reports; the data does not distinguish between these reasons. As shown in the appendix data section, we identify the block-by-block inspections as those for which many other buildings in the same Census tract are inspected in the same week. We provide more details on how we filter out responsive inspections and validate our filtration process in the appendix data section. 

\paragraph{Matching ratings and reports to geographic units:} Next, we match government inspection ratings and crowdsourced reports to geographic regions and to each other. Each rating is for a fine-grained unit within a Census tract, e.g., street ratings are for street segments and rodent ratings are for buildings. For Census tract/type/week tuples with an observed rating, we match the rating to its corresponding fine-grained report. Thus, we may have multiple ratings and reports for each Census tract/type/week tuple. For Census tract/type/week tuples without observed ratings, we aggregate the reporting data at the Census tract level. Thus, we have one rating and report for each Census tract/type/week tuple. See the appendix data section for details.
\section{Model}
\label{sec:model}
\paragraph{Approach overview:} Our model is summarized in Figure \ref{fig:model}. The goals are (i) to estimate the true state of an incident at a location -- what is the true street condition in a neighborhood?; and (ii) to quantify biases in the reporting data -- which neighborhoods systematically underreport incidents and how does reporting correlate with demographics? In many urban settings, models are fit using only the frequently observed reporting data, resulting in biased predictions \citep{xu2017predicting, casey2018cautionary, li2020311, hacker2022spatiotemporal}. In contrast, our approach estimates the true state using both the frequently observed biased reporting data and the sparsely observed unbiased rating data.

\paragraph{Notation:} Consider a network $G$ with $n$ nodes and adjacency matrix $E$. Nodes are indexed by $\node$ and represent neighborhoods, and edges connect adjacent neighborhoods. Each node $\node$ has features $X_\node \in \bbR^{D}$, where $D$ is the number of features. These features include demographic factors that may influence reporting rates. There are $\tau$ total incident types indexed by $k$ (e.g., rodents, floods, etc.). We index time by week $t$. The model uses two types of data: sparsely observed unbiased true state data (e.g., \textit{inspection ratings}) and frequently observed biased data (e.g., \textit{crowdsourced reports}).

\paragraph{Observed data:} We observe inspection ratings $\rikt\in\bbR$ which are unbiased observations of the true latent state, but are only available for a subset of nodes, types, and weeks. We also observe indicators of reports of incidents $\Tikt\in \{0,1\}$, where $\Tikt=1$ indicates that an incident of type $k$ was reported for node $\node$ at week $t$. As discussed in \S\ref{sec:data}, for some types, both inspection ratings and reports are available at the \textit{sub-node} level. We provide the sub-node level notation of our URBAN model in the appendix model section.

\paragraph{Model:} The \textit{true} inspection rating $\rikt$ captures the condition of incident type $k$ in node $\node$ during week $t$. The true ratings are drawn from some distribution $f_r$ as a function of the node, type, and week, as follows:
\begin{align}
\label{eq:rating}
\begin{split}
\text{True inspection rating: } & r_{\node kt} \sim f_r(\cdot|\node, k, t).
\end{split}
\end{align}

We now discuss how we \textit{model} ratings. To model ratings, we learn a \textit{node embedding} and a \textit{type embedding}. Node $\node$'s embedding ${e_n}[i]\in\bbR^{E_{emb}}$, where $E_{emb}$ is the embedding dimension, is a low-dimensional representation of a node and captures the node's attributes and position. The node embeddings are learned using a graph convolutional network (GCN) \citep{kipf2017semi}, which is a model that leverages graph data by iteratively aggregating and transforming features from neighboring nodes. Thus our node embeddings are \textit{spatially correlated}, capturing the correlation of true incident occurrence across neighborhoods. We also learn type $k$'s embedding ${e_\tau}[k]\in\bbR^{E_{emb}}$. The type embedding is a low-dimensional representation of a type and captures the type's features, similarity to other types, and relationship to nodes in the graph. Thus, our type embeddings capture correlations across types. We model ratings as:
\begin{align}
\label{eq:rating}
\begin{split}
\text{Pred. rating: } & \hat{r}_{\node kt} = e_n[\node]^\top e_\tau[k]
\end{split}
\end{align}
In words, the predicted rating $\hat{r}_{\node kt}$ is estimated as the dot product of node $\node$'s embedding and type $k$'s embedding.

We model the probability of observing a report as a type-specific function of demographics and the inspection rating. We learn type-specific reporting parameters to account for the fact that different incident types have different reporting characteristics, a claim we confirm on our real rating data. We model the predicted probability of observing a report $\hat{P}(\Tikt)$ as the following logistic ($\sigma$) function:
\begin{align}
\text{Pred. prob. of report: } & \hat{P}(\Tikt) = \sigma(\alpha^*\rikt^* + \theta^{*\top} X_\node), \label{eq:report}
\end{align}
where $\alpha^*$ is the coefficient on true rating (how relatively likely is an incident with rating $r$ to be reported) and $\theta^*$ is the coefficient on node-specific demographic features (is a node with features $X_i$ more likely to report, given the rating). $\alpha^*$ and $\theta^*$ are interpretable reporting coefficients that allow us to understand how each demographic covariate impacts reporting probabilities.

As we discuss further below, the values used for $\rikt^*,\alpha^*,\theta^*$ depend on whether a rating $\rikt$ is observed in node/type/week tuple $(i,k,t)$ and whether ratings for other nodes $\node'$ or weeks $t'$ for type $k$ are observed ($r_{\node'kt'}$). We must consider these cases separately because we cannot simultaneously learn the rating $\rikt$ and the type-specific reporting coefficients $[\alpha_k, \theta_k]$---there is a fundamental unidentifiability between whether conditions are bad (low rating $\rikt$) but the area does not make a report (small reporting coefficients $[\alpha_k, \theta_k]$) or conditions are good (high rating $\rikt$) and no report is needed. We now describe the three different cases:

\textit{Case 1 -- A rating $\rikt$ is observed:} Here, the probability of observing a report is a function of demographics $X_i$ and the true observed rating, $\rikt^* = \rikt$. We estimate type specific reporting coefficients, $[\alpha_k^*,\theta_k^*] = [\alpha_k,\theta_k]$.

\textit{Case 2 -- No rating $\rikt$ is observed at node $i$ but ratings $r_{\node' kt'}$ for type $k$ are observed at other nodes $\node'$ or times $t'$:} Here, we do not have access to node $\node$'s true rating, so we model the probability of observing a report as a function of demographics $X_i$ and the \textit{predicted ratings}, $\rikt^* = \hat{r}_{\node kt}$. We use type specific reporting coefficients, $[\alpha_k^*,\theta_k^*] = [\alpha_k, \theta_k]$, which are learned via nodes $\node'$ or times $t'$ for which ratings $r_{\node'kt'}$ are observed. 

\textit{Case 3 -- No ratings $\rikt$ for type $k$ are observed at any node or time period:} We do not observe the true rating, so we model the probability of observing a report as a function of demographics $X_i$ and the \textit{predicted ratings}, $\rikt^* = \hat{r}_{\node kt}$. We cannot simultaneously learn the rating $\rikt$ and type-specific reporting coefficients $[\alpha_k, \theta_k]$. Thus, we borrow information from types with observed ratings and use the mean reporting coefficients across these types, $[\alpha_k^*,\theta_k^*] =[\overline{\alpha}, \overline{\theta}]$.

For observed types $k$, we can recover type-specific reporting coefficients $[\alpha_k,\theta_k]$ to account for different types' likelihoods of being reported when incidents occur. For instance, residents may be more likely to report rodents than a noise complaint. We assume that the mean coefficients $[\overline{\alpha}, \overline{\theta}]$ are reasonable for types with unobserved ratings, i.e., the reporting coefficients transfer across types to some extent.

URBAN's modelling contribution lies in adapting GNNs for biased data, and our approach extends to other parameterizations of $\rikt$ and $\Tikt$. While our described model predicts constant ratings $\hat{r}_{\node kt}$ and reporting probabilities $\hat{P}(\Tikt)$ over time, our method generalizes to other GNN architectures, other reporting models, and spatio-temporal methods. We expect all these models \textit{with multiple types of data} to perform well. Regarding time, as new data comes in, our model can be modified to throw away old data (this is how we train models on real data across different time splits), upweight new data, or use a spatiotemporal GNN. Full details are provided in the appendix model section.

\begin{figure*}[t]
    \begin{center}
    \includegraphics[width=0.95\textwidth]{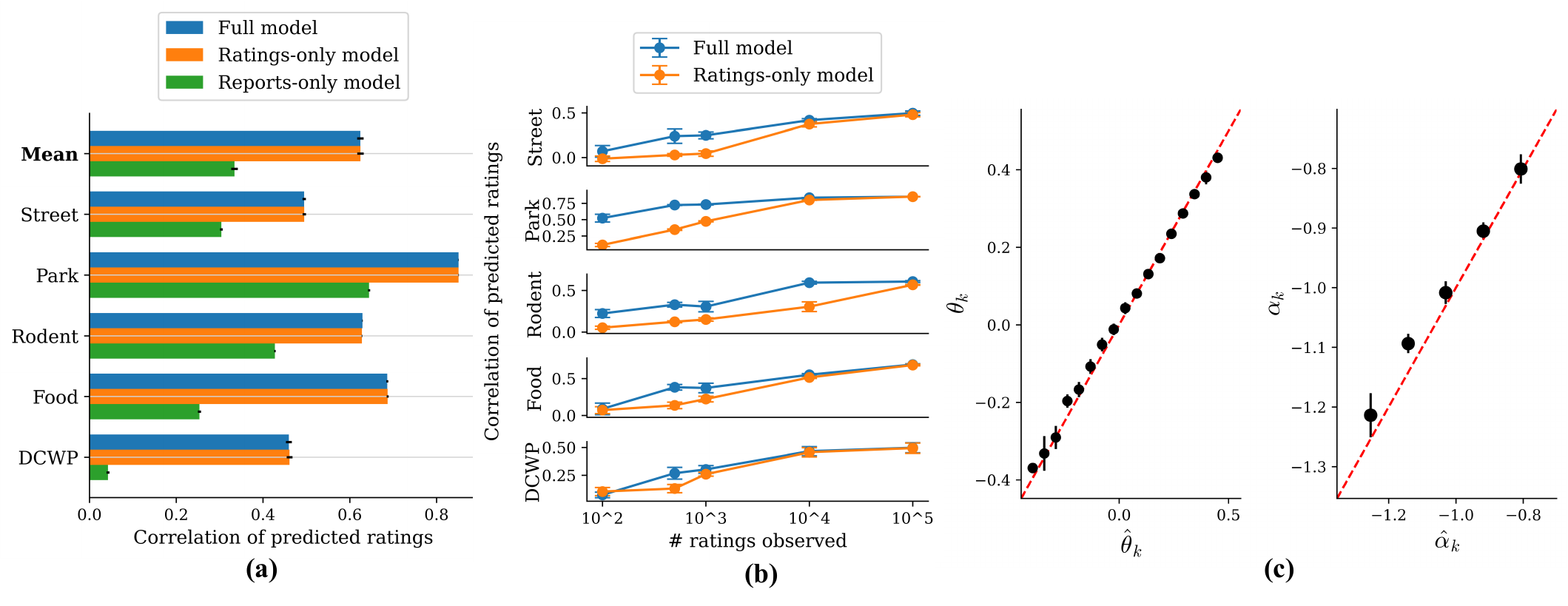}
  \end{center}
    \caption{\textit{Semi-synthetic results.} We calculate the correlation between the average predicted and true rating for each node/type pair. We find that (a) our full URBAN model predicts ratings that are more correlated with ground truth than a model that uses only reporting data, and (b) as ratings are more sparsely observed, for types where reports are predictive of ratings, our full model predicts ratings that are more correlated with ground truth than a model that uses only rating data. (c) Our model's predicted coefficients $[\hat{\theta_k},\hat{\alpha_k}]$ match the true coefficients $[\theta_k,\alpha_k]$ for all types with observed ratings. Panels (a) and (c) show results averaged across 20 semi-synthetic datasets. Panel (b) shows results averaged across 5 datasets.}
    \label{fig:semisynthetic_results}
    \vspace{-1.5em}  
\end{figure*}

\paragraph{Loss function:} Our loss function is a weighted sum of four components that evaluate our URBAN model's performance in predicting reports and ratings:

(i) \textit{Report loss for data points with \textbf{unobserved} ratings:} Binary cross entropy (BCE) loss between the true report status $\Tikt$ and the predicted probability of a report $\hat{P}(\Tikt)$ for data points with unobserved ratings.
\begin{align}
\label{eq:unobserved_report_loss}
\begin{split}
\mathcal{L}_{\textrm{unobs}} = \sum_{ikt} \mathds{1}\left(\rikt \textrm{ unobs}\right) \cdot \textrm{BCE}(\hat{P}(\Tikt), \Tikt)
\end{split}
\end{align}
(ii) \textit{Report loss for data points with \textbf{observed} ratings:} BCE between the true report status and the predicted probability of a report for fine-grained data points with observed ratings.
\begin{align}
\label{eq:observed_report_loss}
\begin{split}
\mathcal{L}_{\textrm{obs}} &= \sum_{ikt} \mathds{1}\left(\rikt \textrm{ obs}\right) \cdot \textrm{BCE}(\hat{P}(\Tikt), \Tikt)
\end{split}
\end{align}
(iii) \textit{Rating loss:} Mean squared error (MSE) between the true rating $\rikt$ and the predicted rating $\hat{r}_{\node kt}$.
\begin{align}
\label{eq:rating_loss}
\begin{split}
\mathcal{L}_{\textrm{rating}} &= \sum_{ikt} \mathds{1}\left(\rikt \textrm{ obs}\right) \cdot \textrm{MSE}(\hat{r}_{\node kt}, \rikt)
\end{split}
\end{align}
(iv) \textit{Regularization loss:} $L^2$ norm of the predicted ratings for data points with unobserved ratings. We include this component to prevent our predicted ratings from growing excessively large. 
\begin{align}
\label{eq:regularization_loss}
\begin{split}
\mathcal{L}_{\textrm{reg}} &= \sum_{ikt}\mathds{1}\left(\rikt \textrm{ unobs}\right) \cdot \hat{r}_{\node kt}^2
\end{split}
\end{align} 
The overall loss is:
\begin{align}
\label{eq:overall_loss}
\begin{split}
\mathcal{L} &= \mathcal{L}_{\textrm{unobs}} + \gamma_1\cdot\mathcal{L}_{\textrm{obs}} + \gamma_2\cdot\mathcal{L}_{\textrm{rating}} + \gamma_3\cdot\mathcal{L}_{\textrm{reg}}
\end{split}
\end{align}
We use weights $\gamma_1, \gamma_2, \gamma_3$ and fix the weight on $\mathcal{L}_{\textrm{unobs}}$ to 1. We select weights via a hyperparameter search minimizing the RMSE of predicted reports and ratings on a validation split. To aid estimation in real data settings we use two additional losses. Full details are in the appendix model section. 
\section{Semi-synthetic experiments}
\label{sec:semisynthetic_experiments}
We now validate our approach on semi-synthetic data, using real reporting data but synthetic ratings. In this setting, our URBAN model is \textit{well-specified} and captures the true relationship between ratings and reports. An advantage of our synthetic data is that we can know and vary the parameters. We verify that in these conditions our model can (i) better predict true ratings compared to using either the ratings or reports alone and (ii) recover the reporting parameters.

\subsection{Experimental setup} 
For our semi-synthetic experiments, we use real reporting data $\Tikt$ as described in the \S\ref{sec:data}, and we use demographic data $X_i$ from the U.S Census Bureau \cite{education, race, income, age, renter, population}. We generate synthetic inspection ratings $\rikt$ by inverting eq. \eqref{eq:report}:
\begin{align}
\label{eq:semisynthetic_rating}
\begin{split}
\rikt &= \frac{1}{\alpha_k}\left(\textrm{logit}(\mathbb{E}_t(\Tikt)) - \theta_k^\top X_\node\right).
\end{split}
\end{align}
$\mathbb{E}_t(\Tikt)$ is the empirical frequency of $\Tikt$ over all weeks in the dataset and $[\alpha_k,\theta_k]$ are the reporting coefficients. Our synthetically generated ratings $\rikt$ aim to distributionally match our real ratings. To match the sparsity pattern, we generate a synthetic version of each rating observed in the real data. To test the limits of our model with sparse data, we also run experiments where we further subsample the observed ratings. To reflect real-world reporting parameters, for each type $k$, we draw reporting parameter $\theta_k$ from a Gaussian with a standard deviation of $0.1$ and a mean equal to the average reporting coefficient predicted by a logistic regression model run on the real rating data.\footnote{We set the intercept and $\alpha_k$ such that our generated $\rikt$ are zero mean and have a standard deviation of 1.} Full details are available in the appendix semi-synthetic experiments section.

We report results averaged across 20 trials. For each trial, we draw a set of coefficients $[\alpha_k,\theta_k]$; generate a new set of ratings; refit the model; and evaluate the predicted ratings, reports, and coefficients. We use a time-based split. We train on data from January 2022 to June 2023 and test on data from July 2023 to December 2023. We assess the effect of using both reporting and rating data by comparing predictions from models using (i) both reporting and rating data (\textit{full model}), (ii) only reporting data (\textit{reports-only model}), and (iii) only rating data (\textit{ratings-only model}).

To evaluate, we compare predictions to true reports and ratings. Our primary metric is the correlation between predicted ratings/probabilities of a report and the average true ratings/report statuses. This measure evaluates how well our model ranks incidents across nodes for each type. the The appendix semi-synthetic experiments section also reports RMSE which behaves similarly. 

\subsection{Semi-synthetic data results} 
Our URBAN model's predicted ratings and reports correlate with ground truth. As shown in the appendix semi-synthetic experiments section, across all types, the average correlation between our full model's predicted probability of a report and the true probability is $r = 0.49$ ($p < 0.001$ for $81\%$ of types). Across all types with observed ratings, the average correlation between our full model's predicted rating and the true rating is $r = 0.62$ ($p < 0.001$ for all types). Our full model recovers the empirical average ratings for each node/type pair in the training data ($r=1.00$) and the remaining error is due to the distribution shift between the train and test time periods. 

\paragraph{The full model outperforms a model that uses only reporting data:} Compared to the reports-only model, the full model predicts ratings that better correlate with ground truth ($r=0.62$ for the full model vs. $0.33$ for the reports-only model). Figure \ref{fig:semisynthetic_results} (a) shows the improvement for each type with observed ratings. The reports-only model's ratings are poorly correlated with the ground-truth ratings because it \textit{cannot} recover the true reporting coefficients.

\begin{figure}
\vspace{-0em}
  \begin{center}
    \includegraphics[width=0.47\textwidth]{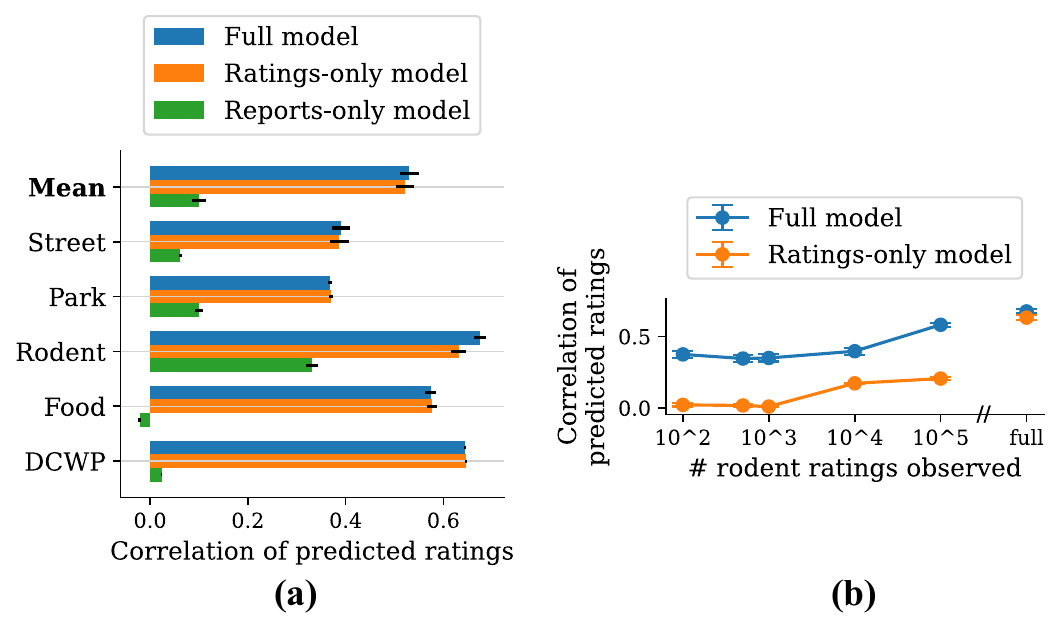}
  \end{center}
  \caption{\textit{Real data results.} We calculate the correlation between the average predicted and true rating for each node/type pair. We find that (a) our full URBAN model predicts ratings that are more correlated with ground truth than a model that uses only reporting data, and (b) as ratings are more sparsely observed, for the rodent type where reports are predictive of ratings, our full model predicts ratings that are more correlated with ground truth than a model that uses only rating data. We plot the mean correlation and 95\% CIs over contiguous two-year periods between 2021 and 2023.}
  \label{fig:real_results}
  \vspace{-1.5em}
\end{figure}

\paragraph{The full model outperforms a model that only uses rating data when rating data is sparse:} Next we compare our full model to the ratings-only model. We show that the full model is valuable when ratings are sparsely observed. We run experiments with artificial sparsity by subsampling ratings for each type with observed ratings. We plot the results in Figure \ref{fig:semisynthetic_results} (b). For types where reports are predictive of ratings (reports-only model achieves a high correlation, e.g. park and rodent), as we decrease the number of observed ratings, the full model relies on reporting data and the predicted ratings remain correlated with the true ratings, while the ratings-only model approaches random prediction ($r=0$). More details on the subsampled experiments are in the appendix model section.

\paragraph{The model recovers the true reporting coefficients:} As shown in Figure \ref{fig:semisynthetic_results} (c), our model recovers the true reporting coefficients $[\alpha_k,\theta_k]$ for incident types with observed ratings. This can only be done by using both types of data together. With only reporting data it is impossible to distinguish between a bad inspection rating that is never reported and a truly good inspection rating. With only rating data we cannot estimate reporting coefficients. Thus, to identify reporting coefficients, one must either use \textit{both} rating and reporting data or make strong parametric assumptions. Additionally, our model only estimates type-specific reporting coefficients for incident types with observed ratings and we accordingly evaluate its performance on these types. For incident types without observed ratings, we use the average reporting coefficients learned across incident types with observed ratings.

\paragraph{The model can learn across types:} In the appendix semi-synthetic experiments section we show that, compared to the reports-only model, our full model predicts ratings that are more correlated with ground truth even for types with fully unobserved ratings. Thus rating data for \textit{observed} types helps in predicting ratings for \textit{unobserved} types, and our model can generalize \textit{across} types.

Overall, our results show that our approach helps if our URBAN model is well-specified. In the next section, we assess on real data.
\section{Real-world case study: New York City crowdsourced reporting} 
\label{sec:real_data_experiments}
We now evaluate our approach using real data, both in terms of predicting ratings and in recovering reporting biases. 

\subsection{Experimental setup}
\label{sec:experimental_setup}
We use real reporting data $\Tikt$ and rating data $\rikt$, as described in \S\ref{sec:data}. As in the semi-synthetic experiments, we use demographic data $X_i$ from the U.S Census Bureau \cite{education, race, income, age, renter, population}. When a government inspection occurs, we assume that the resulting rating remains constant over time until the next inspection occurs. We split our data into a train and test set using a time-based split, as is standard in urban prediction tasks \citep{yu2018spatiotemporal, farahmand2023spatial, huang2023temporal, agostini2024bayesian}. We create splits over every contiguous two-year period between January 2021 and December 2023 (13 total periods). For each two year time period, we train on the first 18 months of data and test on the last 6 months. We report results averaged across all 13 time splits.\footnote{We report CIs over the 13 overlapping two-year splits between Jan 2021 and Dec 2023.} 

\subsection{Results}
\label{sec:validating_the_model}
Prediction on real data is more challenging than semi-synthetic data due to model misspecification. We model the probability of a report as a logistic function of demographics and ratings, which allows us to quantify how reporting rates vary by demographics. But in reality, it is likely that reports are generated by a more complex function with more complex inputs. Nevertheless, our model's predicted ratings and reports still correlate with ground truth. Across all types, the average correlation between our full model's predicted probability of a report and the true probability of a report is $r = 0.24$ ($p<0.001$ for $83\%$ of types). As shown in the appendix real data experiments section, across all types with observed ratings, the average correlation between our full model's predicted rating and the true rating is $r = 0.53$ ($p<0.001$ for all types). Additionally, in the appendix real data experiments section we report RMSE results; we evaluate rating predictions using top-$k$ coverage, expected calibration error (ECE), correlation and ECE gaps for demographic subgroups, and spatial equity under budget constraints as measured by the average demographics of Census tracts which receive resources compared to the overall demographics; and we discuss performance across Census tracts.

\paragraph{The full model outperforms a model that uses only reporting data:} The correlation between the full model's predicted ratings and the ground-truth ratings is $r=0.53$ vs. $0.10$ for the reports-only model. Figure \ref{fig:real_results} (a) shows the full model's improvement in predicting ratings for each type with observed ratings. We also compare to Gaussian process (GP) models trained only on the reporting data. We separately fit models for each type with observed ratings. The GP ($r = 0.03$) underperforms both the full model and the reports-only model showing the value of a flexible model that learns across multiple incident types and demographics. 

Our key finding is that any model trained on only one source of data is limited. Even test split reports are a poor predictor of test split ratings ($r=0.0180$ vs. $0.5303$ for our full model). Prior work has also found that reports alone cannot accurately predict ground truth \cite{casey2018cautionary}. This confirms that there is a low ceiling to the performance of any ML method trained only on reports.

\paragraph{The full model outperforms a model that only uses rating data when rating data is sparse:} Next we compare our full model to the ratings-only model. We show that the full model is valuable when ratings are sparsely observed and reports are predictive of ratings. We run experiments with artificial sparsity by subsampling observed ratings. As shown in Figure \ref{fig:real_results} (b), for the rodent type, where reports are predictive of ratings, as we decrease the number of observed ratings, our full model's predicted ratings remain correlated with the true ratings, while the ratings-only model approaches random prediction ($r=0$). For the other types with observed ratings, reports are not predictive of ratings (reports-only model achieves a low correlation, $r \leq 0.1$), the model's predictive power comes from rating data, and reporting data minimally improves rating predictions.

\begin{table}[]
\vspace{-0em}
\centering
\begin{tabular}{l r}
\hline
\textbf{Covariate}               & \textbf{Mean coefficient} \\ \hline
log(Population density)             & $0.250\pm0.058$                            \\
log(Median income)                      & $0.173\pm0.020$                           \\
Bachelors degree population       & $0.159\pm0.018$\\
Households occupied by renter                 & $0.115\pm0.029$                            \\
Median age                         & $0.104\pm0.016$                            \\
White population                   & $0.093\pm0.012$                           \\
True inspection rating                        & $-0.197\pm0.009$                            \\
\hline
\end{tabular}
\caption{\textit{Demographic coefficients.} We report the average predicted univariate coefficients across types with observed ratings. The estimated coefficients capture known demographic factors: e.g., tracts that have a higher income are more likely to report incidents. We also report the average coefficient on the true inspection rating across all models. Tracts that have worse ratings are more likely to have reports. We report the mean coefficients and 95\% CIs over all contiguous two year periods between 2021 and 2023.}
\vspace{-2em}
\label{tab:real_coefficients}
\end{table}

\paragraph{The model learns demographic predictors of underreporting:} $\theta_k$ measures the contribution of each demographic feature in $X_\node$ to the reporting rate. We estimate $\theta_k$ by fitting univariate variants of our full model. Each univariate model is trained on both rating and reporting data but only uses one demographic feature. We run a separate univariate model for each demographic feature in $X_\node$. Table \ref{tab:real_coefficients} shows that the inferred coefficients capture known demographic predictors of underreporting. We find that conditional on ground-truth, neighborhoods with higher population density, higher median income, higher proportion of the population with a bachelor's degree, older residents, higher proportion of white residents, and worse ground-truth all report incidents at higher rates; these estimates are consistent with those of other work \citep{kontokosta2021bias, agostini2024bayesian, liu2024quantifying,matt2025flooding} and add further evidence of demographic bias in crowdsourced reporting. Coefficients estimated by a multivariate model are in the appendix real data experiments section.

\paragraph{Our full model's predicted ratings are spatially correlated and capture correlations across incident types:} We evaluate the interpretability of our model's predicted ratings by analyzing how well they correlate across nodes and types. We first analyze the predicted ratings' spatial correlation. For each node $\node$, we create a vector $\mathbf{r}_i=\{\rikt\}_{k=1}^{\tau}$ of predicted ratings over all types $k$. We use each node's $\mathbf{r}_i$ vector to cluster the nodes into $4$ groups. As shown in the appendix real data experiments section, we find that our clusters are spatially correlated demographically different. We also show that our full model learns more spatially correlated ratings than the ratings-only model. The higher spatial correlations from our full model are consistent with prior work \citep{agostini2024bayesian} and potentially rendes our full model to be more interpretable.

Next, we analyze our predicted ratings across complaint types. For each type $k$, we create a vector $\mathbf{r}_k=\{\rikt\}_{i=1}^{n}$ of predicted ratings over all nodes and cluster the types into $8$ groups. Each group contains a coherent cluster of types which we describe in the appendix real data experiments section. Additionally, we show that the dimension of highest variability of $\mathbf{r}_k$ captures type frequency (i.e., $\mathbb{E}_{it}[\Tikt]$). These results show that both our predicted and true ratings are spatially correlated and capture similarities across types.
\section{Discussion}
\label{sec:discussion}
We address the challenging problem of estimating graph neural networks (GNNs) in settings where we observe biased outcome data. In these settings, nodes have a true latent state that is sparsely observed (e.g., via government inspection ratings). We also frequently observe biased proxies of the latent states (e.g., via crowdsourced reports). We propose a GNN-based model, URBAN, that jointly uses frequently observed biased reporting data and sparsely observed unbiased rating data. We apply our model to New York City 311-data and show that (i) our model makes better predictions of the ground-truth latent state compared to both a baseline model trained only on reporting data and a baseline model trained only on rating data; (ii) our model's inferred reporting coefficients capture known demographic factors associated with underreporting; and (iii) our model's learned ratings are interpretable and capture correlation between nodes and 311-complaint types.

Our findings have important implications for urban resource allocation decisions. NYC, like many municipalities, uses 311 data to help schedule inspections \cite{rodent_ratings, tussey2025principles}. Prior work has shown that learning \textit{only} from reporting data can lead to biased predictions and downstream inequities in resource allocation. For example, \citet{casey2018cautionary} found that in D.C., reports on rodents did not accurately predict ground truth inspection outcomes. Our results reinforce that decisions based \textit{exclusively} on reporting data risk perpetuating demographic biases and misallocating resources, and that jointly modeling reporting data with ground truth outcomes can aid prediction. 

There are several avenues for future work. Our model is an example of how to estimate reporting propensity, but future work could investigate whether other methods of incorporating reporting data could produce more accurate rating predictions. Another direction is to apply our model to other settings where demographic reporting disparities exist, such as, crime reporting, public health incident tracking, and across domains where equitable resource allocation depends on accurate understanding of underlying conditions.

\section{Acknowledgments}
The authors thank Vince Bartle, Serina Chang, Erica Chiang, Matt Franchi, Sophie Greenwood, Raj Movva, and Kenny Peng for helpful conversations. SB's work is supported by NSF GRFP Grant DGE \#2139899. SS's work is supported by NSF GRFP Grant DGE \#2141064 and the Hertz Fellowship. EP's work is supported by a Google Research Scholar award, an AI2050 Early Career Fellowship, NSF CAREER \#2142419, a CIFAR Azrieli Global scholarship, a gift to the LinkedIn-Cornell Bowers CIS Strategic Partnership, the Survival and Flourishing Fund, Open Philanthropy, and the Zhang Family Endowed professorship. NG's work is supported by NSF CAREER IIS-2339427, NASA, the Sloan Foundation, and Cornell Tech Urban Tech Hub, Google, Meta, and Amazon research awards. Any opinions, findings, conclusions, or recommendations expressed in this material are those of the authors and do not necessarily reflect the views of the funders.

\bibliography{aaai2026}

\newpage
\appendix
\begin{table*}[h]
    \centering
\begin{tabularx}{\textwidth}{>{\raggedright}Xccccc} 
    \toprule[1.5pt]
    & \textbf{Street} & \textbf{Park} & \textbf{Rodent} & \textbf{Food} & \textbf{DCWP} \\
    \midrule[1.5pt]
    \textbf{Number of crowdsourced reports} & 217k & 56k & 122k & 48k & 13k \\ 
    \hline
     \textbf{Number of government inspection ratings} & 91k & 18k & 238k & 30k & 9k \\ 
     \hline
    \textbf{Fine-grained collection unit} & street segment & park & BBL & BBL & Census block \\ 
    \hline
     \textbf{How are ratings and reports matched?} & Matching & Matching & Available in data & Available in data & Available in data \\ 
    \bottomrule[1.5pt]
\end{tabularx}
    \caption{\textit{Data summary.} For each of the five incident types for which we process real ratings, we report the total number of reports made between 2021 and 2023, inclusive; the total number of ratings collected between 2021 and 2023, inclusive; what fine-grained unit ratings are collected at (e.g., Borough-Block-Lot or BBL); and whether corresponding ratings and reports were matched by a matching heuristic (each rating was mapped to its nearest report) or whether the matching was available in the data.}
    \vspace{-1.5em}
    \label{tab:data_summary}
    \end{table*}

\section{Further details on related work}
\label{sec:appendix_related_work}
We discuss in more detail the three areas of literature related to our work.

\paragraph{Predicting urban incidents:} A large literature seeks to predict urban incidents, such as flooding, infrastructure quality, and crime \citep{budde2014leveraging, o2015ecometrics, lum2016predict, sarto2016bayesian, wang2017structure, hu2018urban, mosavi2018flood, yu2018spatiotemporal, gomez2019spatiotemporal, pan2019matrix, xie2019sequential, clark2020advanced, jimenez2020rapid, kontokosta2021bias, agonafir2022understanding, hacker2022spatiotemporal, mauerman2022high, farahmand2023spatial, zaouche2023bayesian, agostini2024bayesian, gao2024improve}. A challenge is that data is biased often due to human behavior \citep{lum2016predict, wang2017structure, clark2020advanced, kontokosta2021bias, agostini2024bayesian, gao2024improve, liu2024quantifying}. With relevance to our setting, prior work has shown that crowdsourced reporting data may not represent ground-truth states due to selection bias via underreporting \citep{wang2017structure, clark2020advanced, kontokosta2021bias, agostini2024bayesian, gao2024improve, liu2024quantifying}.  For example, prior work on urban incident crowdsourcing has quantified underreporting of floods using spatiotemporal models \citep{agostini2024bayesian} and has quantified the geographic and demographic patterns of underreporting \citep{wang2017structure, clark2020advanced, kontokosta2021bias, gao2024improve, liu2024quantifying}. Understanding the patterns of underreporting is crucial because disparities in incident reporting rates lead to downstream inequities in resource allocation \citep{liu2024quantifying, liu2024redesigning}. Especially notable in relation to our work is that of \citet{casey2018cautionary}, who show in Washington, D.C., that 311-reports alone are poor predictors of ground-truth ratings.

\paragraph{Using GNNs for biased data:} A common approach to addressing selection bias in GNNs is \textit{graph-based semi-supervised learning}, which uses two primary methods to infer labels for unlabeled datapoints \cite{song2023graph}: graph regularization, which propagates labels from labeled to unlabeled nodes using the graph structure \citep{zhou2003learning, zhu2003semi, belkin2006manifold, dai2021nrgnn, verma2021graphmix, li2023informative, qian2023robust, zhou2024varianceenlarged}, and graph embedding, which generates low-dimensional representations for nodes to infer labels for unlabeled nodes \citep{perozzi2014deepwalk, tang2015line, grover2016node2vec, li2021comatch, sharma2023efficiently}. Other recent works have generated pseudo-labels using softmax predictions \citep{arazo2020pseudo}, curriculum learning \citep{zhang2021flexmatch}, and a combination of entropy based filtering and contrastive learning \citep{wang2022semi}. Other GNN approaches to biased data include causal regularization \citep{kyono2020castle, wang2022causalgnn, wu2022causally} and graph attention \citep{wang2019improving, brody2022how, sui2022causal}.

\paragraph{Using GNNs for spatiotemporal modeling:} GNNs are a natural fit for high-dimensional spatiotemporal modeling in applications like traffic forecasting, epidemic forecasting, and molecular dynamics \citep{kapoor2020examining, roy2021sst, wang2022causalgnn, wang2022traffic, he2023stgc, pineda2023geometric, wu2024equivariant}. Several works also design ways to encode spatiotemporal information in GNNs, including kriging convolutional networks \citep{appleby2020kriging}, positional encoders \citep{klemmer2023positional}, and inductive kriging \citep{wu2021inductive}. Non-GNN-based spatiotemporal modeling, including Bayesian models, clustering, and matrix factorization models, have also been used for urban issues like infrastructure monitoring \citep{budde2014leveraging}, crime \citep{hu2018urban}, urban flow \citep{pan2019matrix}, air pollution \citep{sarto2016bayesian}, and pedestrian traffic \citep{zaouche2023bayesian}.

Our study also relates to literature on \textit{multi-view GNNs}, which integrate information from multiple datasets or views to improve representation learning and predictive performance \citep{shao2023heterogenous, cheng2018spatiotemporal}, and \textit{multi-output GNNs}, which predict multiple outputs from a single graph by leveraging shared graph structure and correlations between the target outputs \citep{peng2022multi, samal2022multioutput}. 
\section{Further details on the data}
\label{sec:appendix_real_data}
\paragraph{Details on processing real reporting data:} We use reports $\Tikt$ from New York City 311-data \citep{311data}. We analyze all Census tracts with valid demographic information ($n=2292$ nodes), complaint types with a reporting frequency greater than 0.1\% ($\tau=139$ types), and all weeks in the three years from 2021-2023. $\Tikt \in \{0, 1\}$ denotes whether at least one report of type $k$ was made in node $\node$ during week $t$. In total we analyze 9,615,863 reports. 

\paragraph{Feature processing:} 
We include demographic features collected for each Census tract. The full list of features that we include is: log population density, percentage of population with a bachelors degree, percentage of households occupied by a renter, log median income, percentage of population that is white, and median age \cite{education, race, income, age, renter, population}. We normalize all features to have mean 0 and standard deviation 1.

\paragraph{Details on processing real rating data:} We collect ratings from government inspection data for five complaint types: (i) street conditions \citep{street_ratings}, (ii) park maintenance or facility conditions \citep{park_ratings}, (iii) rodents \citep{rodent_ratings}, (iv) food establishment/mobile food vendor/food poisoning \citep{restaurant_ratings}, and (v) DCWP consumer complaints \citep{dcwp_ratings}. A summary of the rating data across these five types is provided in Table \ref{tab:data_summary}. Each rating is for a fine-grained unit within a Census tract. Street ratings are for street segments; park ratings are for parks; rodent and food ratings are averaged over each Borough-Block-Lot (BBL); and DCWP ratings are averaged over each Census block. We match each fine-grained rating to its corresponding fine-grained report (i.e., reports in that same street segment). For rodents, food, and DCWP the matching is done directly (i.e., we match the aggregated rodent rating for a BBL to the aggregated report for the same BBL). For streets and parks, to account for potential error in the latitude/longitude information for each 311-report, we run a distance heuristic to complete the matching. We match each rating with its nearest report. If the nearest report is above a certain distance threshold, we filter out the rating (consider it unobserved). Within the same tract, all fine-grained ratings and reports are provided to the model and are mapped to the same node's embedding, as well as the corresponding type's embedding.

\begin{figure}
\vspace{-0em}
  \begin{center}
    \includegraphics[width=0.45\textwidth]{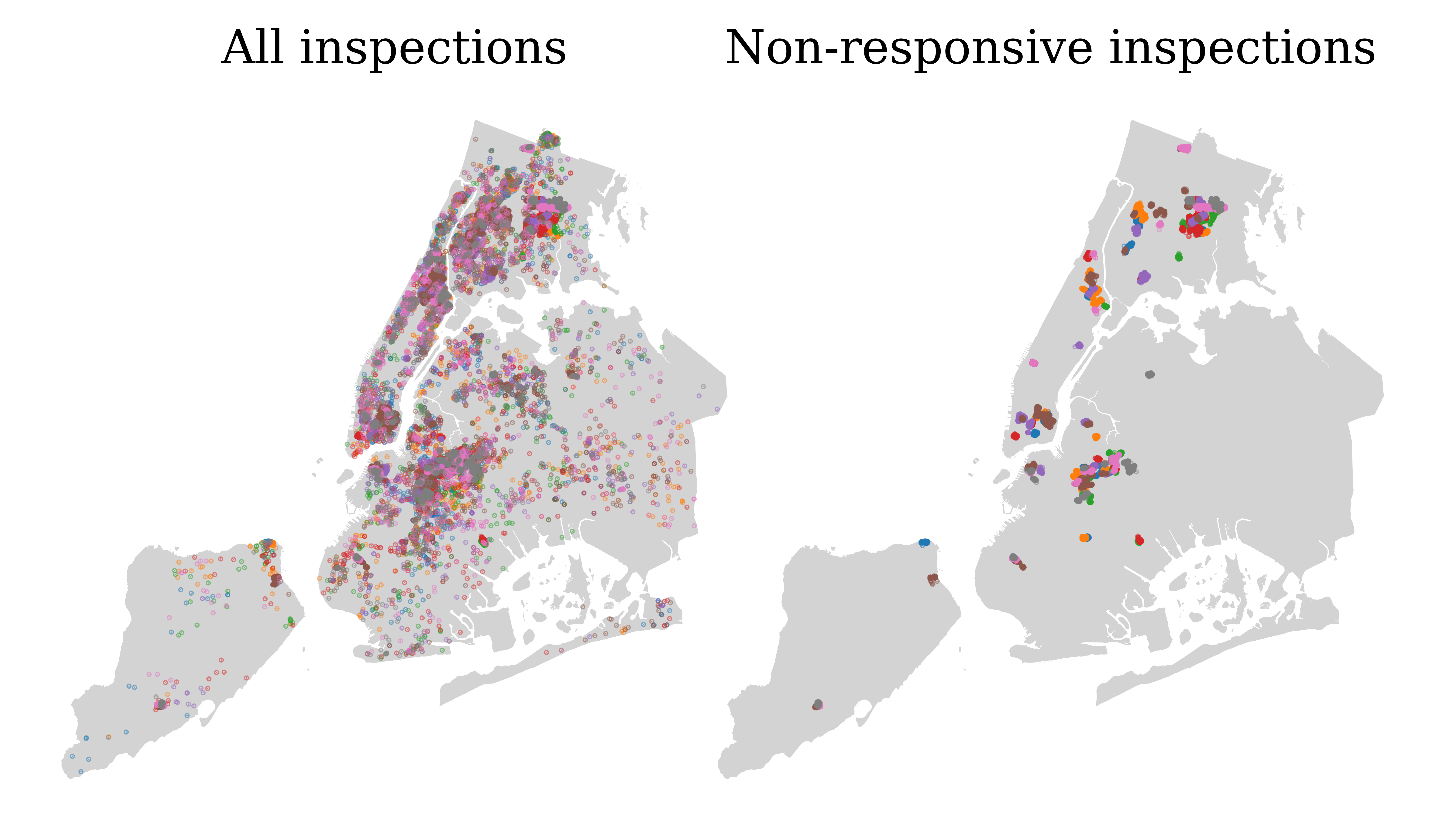}
  \end{center}
  \caption{\textit{Data processing.} We process the rodent inspections to remove all inspections triggered by crowdsourced reports. This figure shows the location of rodent inspections for the first 8 weeks of 2023. While the responsive inspections are distributed throughout the city (left), after we apply a heuristic, we can identify only the non-responsive inspections (right). The non-responsive inspections are clustered together because these inspections are conducted block-by-block in a scheduled manner \citep{rodent_ratings}.}
  \label{fig:random_inspections}
  \vspace{-1.5em}
\end{figure}

In order for our rating data to represent \textit{unbiased} observations of the true incident states, we must process the data to ensure the inspections are conducted proactively. In particular, we remove inspections triggered by 311-reports. For example, each week DOHMH rodent inspectors run two types of inspections: (i) inspections triggered by 311-reports and (ii) inspections on multiple buildings in a randomly selected set of blocks \citep{rodent_ratings}. Since the data does not indicate what type of inspection was run, we create a heuristic to identify the block by block inspections. Our heuristic selects for inspections that are clustered close together (i.e., in the same block). In particular, we calculate the number of inspections that occur each week in each Census tract, and we keep all inspections that fall in tracts at least at the 50th percentile. Figure \ref{fig:random_inspections} shows that our heuristic picks out the proactively run block by block rodent inspections. To validate our heuristic, we ensure that: (i) Ratings are lower for filtered inspections (the average rating for filtered inspections is 1.1 while the average rating for unfiltered ratings is 1.6). (ii) For a given block, ratings are lower during weeks with reactive inspections (the average rating for weeks with reactive inspections is 1.2 while the average rating for weeks without reactive inspections is 1.6). (iii) There is less time between a report and inspection for filtered inspections (the average amount of time between a report and inspection for filtered inspections is 5.8 weeks while the average amount of time between a report and inspection for unfiltered inspections is 13.0 weeks). Thus unfiltered inspections are \textit{less} likely to be responsive and are not skewed to worse ratings. We note that the ratings for the other four types we process (street, park, restaurant, and DCWP) are not made in response to reports and thus we do not need to process them further. We also note that while our approach produces ratings that do not present any reporting biases, the ratings could display other biases, e.g. inter-inspector variability \citep{zamfirescu2022trucks}.
\section{Further details on the URBAN model}
\label{sec:appendix_model}
Here we describe our model using notation for the \textit{sub-node} level that some observed ratings and reports are collected at. We also provide additional details about our model and learning procedure.

\paragraph{Notation:} We use the same notation as \S\ref{sec:model}. Nodes are indexed by $\node$ and represent neighborhoods. Incident types are indexed by $k$ (e.g., rodents, floods, etc.). We index time by week $t$. We introduce the following notation: Each node $i$ contains fine-grained geographic units indexed by $i^*$ (e.g. each Census tract contains street segments or buildings).

\paragraph{Observed data:} We observe inspection ratings at the sub-node level. Thus, for some fine-grained geographic unit/type/week tuples, we observe ratings $r_{i^*kt}\in\bbR$. We also observe indicators of reports of incidents. For node/type/week tuples where a ratings is observed, indicators of reports $T_{i^*kt}\in \{0,1\}$ are also observed at the sub-node level. For node/type/week tuples where \textit{no} rating is observed, indicators of reports $\Tikt\in \{0,1\}$ are observed at the node level.

\paragraph{Model:} 
The \textit{true} inspection rating $r_{i^*kt}$ captures the condition of incident type $k$ in fine-grained unit $i^*$ during week $t$. More formally, the true ratings are drawn from some distribution $f_r$ as a function of the node, type, and week, as follows:
\begin{align}
\label{eq:rating_finegrained}
\begin{split}
\text{True inspection rating: } & r_{\node^*kt} \sim f_r(\cdot|\node, k, t).
\end{split}
\end{align}

We model ratings at the node level using learned node and type embeddings as follows:
\begin{align*}
\begin{split}
\text{Predicted rating: } & \hat{r}_{\node kt} = e_n[\node]^\top e_\tau[k]
\end{split}
\end{align*}
In words, the predicted rating $\hat{r}_{\node kt}$ is estimated as the dot product of node $\node$'s embedding $e_n[\node]$ and type $k$'s embedding $e_\tau[k]$. Note that our model predicts constant ratings across each fine-grained geographic unit $i^*$ in node $i$. 

We model the probability of observing a report as a type specific logistic function $(\sigma)$ of the inspection rating and demographics. Since the true inspection ratings are only observed in a subset of node/type/week tuples $(i,k,t)$, we learn a different reporting model depending on whether a rating $r_{i^*kt}$ is observed in node $i$ and whether ratings in other nodes $\node'$ for type $k$ are observed (i.e., whether ratings $r_{\node'^*kt}$ are observed). Additionally, since ratings are observed at the sub-node level, for node/type/week tuples $(i,k,t)$ in which ratings are observed, we learn the probability of observing a report at the sub-node level as well (i.e., we learn $P(T_{i^*kt})$). Overall, there are three different cases that we consider.

\textit{Case 1 -- A rating $r_{i^*kt}$ is observed:} In this case, we model the probability of observing a report as a function of the true granular observed rating $r_{i^*kt}$ and demographic features for the corresponding node $X_i$. We estimate type specific reporting coefficients $[\alpha_k,\theta_k]$. The observed rating is at the sub-node level, and therefore we model the probability of observing a report at the sub-node level as well.
\begin{align}
\label{eq:predicted_report_observed_rating}
\begin{split}
\text{Case 1: } &\hat{P}(T_{i^*kt}) = \sigma(\alpha_kr_{i^*kt} + \theta_k^\top X_\node)
\end{split}
\end{align}
\textit{Case 2 -- No rating $r_{i^*kt}$ is observed in node $i$ but ratings $r_{\node'^* kt}$ for type $k$ are observed in other nodes $\node'$:} In this case, we do not have access to true ratings in node $\node$, so we model the probability of observing a report as a function of the \textit{predicted rating} $\hat{r}_{\node kt}$ and demographic features $X_i$. We use type specific reporting coefficients $[\alpha_k, \theta_k]$ which are learned via eq. \eqref{eq:predicted_report_observed_rating} using the nodes $\node'$ for which the ground truth ratings $r_{\node'^* kt}$ are observed for type $k$. The predicted rating is at the node level, and therefore we model the probability of observing a report at the node level as well.
\begin{align}
\label{eq:predicted_report_observed_type}
\begin{split}
\text{Case 2: } &\hat{P}(\Tikt) = \sigma(\alpha_k\hat{r}_{\node kt} + \theta_k^\top X_\node)
\end{split}
\end{align}
\textit{Case 3 -- No rating $r_{i^*kt}$ is observed for type $k$ at any node:} We again do not have access to the true rating, so we model the probability of observing a report as a function of the \textit{predicted rating} $\hat{r}_{\node kt}$ and demographic features $X_i$. We cannot simultaneously learn the rating $\rikt$ and the type specific reporting coefficients $[\alpha_k, \theta_k]$, thus in this case we model the probability of observing a report as a function of the mean reporting coefficients across types with observed ratings $[\overline{\alpha}, \overline{\theta}]$. The predicted rating is at the node level, and therefore we model the probability of observing a report at the node level as well.
\begin{align}
\label{eq:predicted_report_unobserved_type}
\begin{split}
\text{Case 3: } &\hat{P}(\Tikt) = \sigma(\overline{\alpha}\hat{r}_{\node kt} + \overline{\theta}^\top X_\node)
\end{split}
\end{align}

\begin{table*}[]
    \centering
    \renewcommand{\arraystretch}{1.5} 
    \vspace{-0em}
    \begin{tabularx}{\textwidth}{l
                                 >{\centering\arraybackslash}X 
                                 >{\centering\arraybackslash}X 
                                 >{\centering\arraybackslash}X 
                                 >{\centering\arraybackslash}X}
    \toprule[1.5pt]
     & & \textbf{Full model} & \textbf{Ratings-only model} & \textbf{Reports-only model} \\ 
    \midrule[1.5pt]
   \multirow{4}{*}{\rotatebox[origin=c]{90}{\textbf{\rule{0pt}{15pt}Semi-synthetic\rule{0pt}{15pt}}}} 
   &  \textbf{Correlation of reports} & $0.4680 \pm 0.0020$ & -- & $0.4840 \pm 0.0007$ \\ 
   & \textbf{RMSE of reports} & $0.0774 \pm 0.0009$ & -- & $0.0586 \pm 0.0001$ \\ 
    \cmidrule(l{0pt}r{0pt}){2-5}
    & \textbf{Correlation of ratings} & $0.6243 \pm 0.0075$ & $0.6246 \pm 0.0075$ & $0.3345 \pm 0.0077$ \\ 
    & \textbf{RMSE of ratings} & $1.0358 \pm 0.0257$ & $1.0331 \pm 0.0262$ & -- \\ 
    \midrule[1.5pt]
    \multirow{4}{*}{\rotatebox[origin=c]{90}{\textbf{\rule{0pt}{15pt}Real data\rule{0pt}{15pt}}}} 
    & \textbf{Correlation of reports} & $0.2371 \pm 0.0110$ & -- & $0.5406 \pm 0.0063$ \\ 
    & \textbf{RMSE of reports} & $0.1104 \pm 0.0025$ & -- & $0.0566 \pm 0.0012$ \\ 
    \cmidrule(l{0pt}r{0pt}){2-5} 
    & \textbf{Correlation of ratings} & $0.5303 \pm 0.0194$ & $0.5223 \pm 0.0185$ & $0.0993 \pm 0.0143$ \\ 
    & \textbf{RMSE of ratings} & $0.5833 \pm 0.0123$ & $0.5852 \pm 0.0116$ & -- \\ 
    \bottomrule[1.5pt]
    \end{tabularx}
    \caption{\textit{Experimental results.} For both semi-synthetic and real data, we compare our full URBAN model (which uses both reporting and rating data) to a reports-only and a ratings-only model. Compared to both baselines, our full model can estimate ratings without compromising accuracy in predicting reports. We calculate the correlation and RMSE between our predicted probabilities of reports and the true probabilities for all node/type pairs. We calculate the correlation and RMSE between our predicted ratings and the true ratings for all nodes and for all types with observed ratings. On the semi-synthetic data, we report the mean and 95\% CIs across 20 synthetic datasets. On the real data, we report the mean and 95\% CIs over all contiguous two year periods between 2021 and 2023.}
    \vspace{-1.5em}
    \label{tab:results}
\end{table*}

\paragraph{Embeddings:} We learn node embeddings using a graph neural network (GNN) \citep{kipf2017semi, velickovic2018graph}, which is a deep learning model that leverages graph-structured data by iteratively aggregating and transforming feature information from neighboring nodes. The GNN takes in as inputs the graph $G$ and a set of features for each node $\node$. We use one-hot node features which is common in settings like ours without natural node features \citep{cui2022on}. We learn the type embeddings using a linear layer with one-hot type feature inputs.

The details of our URBAN model architecture are as follows: We use a 2 layer GNN where each layer consists of a graph convolution, leaky ReLU activation, and batch normalization. We use an intermediate dimension equal to the number of nodes $n=2292$ and an embedding dimension of $E_n=E_\tau=50$.

\paragraph{Batching:} We batch our data such that data points with observed ratings and unobserved ratings are batched separately. During training, we freeze the reporting model for batches for which there are no observed ratings (i.e., we learn the reporting coefficients only from types for which ratings are observed).

\paragraph{Loss function:} We calculate loss on both our predicted ratings and our predicted probabilities of reports. For node/type/week tuples $(\node, k,t)$ with observed ratings, we observe fine-grained rating and reporting data (e.g., for each individual street segment), and thus our model also calculates loss at the fine-grained level.

Our final loss function is a weighted sum of four loss components:

(i) \textit{Report loss for data points with \textbf{unobserved} inspection ratings:} Binary cross entropy (BCE) between the true report status $\Tikt$ and predicted probability of a report $\hat{P}(\Tikt)$ for data points with unobserved inspection ratings.
\begin{align}
\label{eq:unobserved_report_loss}
\begin{split}
\mathcal{L}_{\textrm{unobs}} &= \sum_{ikt} \mathds{1}\left(\rikt \textrm{ is unobserved}\right) \cdot \textrm{BCE}(\hat{P}(\Tikt), \Tikt)
\end{split}
\end{align}
(ii) \textit{Report loss for data points with \textbf{observed} inspection ratings:} BCE between the true fine-grained report status $T_{i^*kt}$ and the predicted fine-grained probability of a report $\hat{P}(T_{i^*kt})$ for data points with fine-grained observed inspection ratings.
\begin{align}
\label{eq:observed_report_loss}
\begin{split}
\mathcal{L}_{\textrm{obs}} &= \sum_{i^*kt} \mathds{1}\left(r_{\node^*kt} \textrm{ is observed}\right) \cdot \textrm{BCE}(\hat{P}(T_{i^*kt}), T_{i^*kt})
\end{split}
\end{align}
(iii) \textit{Rating loss:} Mean squared error (MSE) between the true fine-grained rating $r_{\node^* kt}$ and the predicted fine-grained rating $\hat{r}_{\node^* kt}$.
\begin{align}
\label{eq:rating_loss}
\begin{split}
\mathcal{L}_{\textrm{rating}} &= \sum_{i^*kt} \mathds{1}\left(r_{\node^*kt} \textrm{ is observed}\right) \cdot \textrm{MSE}(\hat{r}_{\node^* kt}, r_{\node^* kt})
\end{split}
\end{align}
(iv) \textit{Regularization loss:} $L^2$ norm of the predicted ratings for data points with unobserved ratings. We include this loss to maintain stable training and prevent our predicted ratings from growing excessively large. 
\begin{align}
\label{eq:regularization_loss}
\begin{split}
\mathcal{L}_{\textrm{reg}} &= \sum_{ikt}\mathds{1}\left(r_{\node kt} \textrm{ is unobserved}\right)\cdot L^2(\hat{r}_{\node kt})
\end{split}
\end{align} 
The overall loss is as follows:
\begin{align}
\label{eq:overall_loss}
\begin{split}
\mathcal{L} &= \mathcal{L}_{\textrm{unobs}} + \gamma_1\cdot\mathcal{L}_{\textrm{obs}}+ \gamma_2\cdot\mathcal{L}_{\textrm{rating}} + \gamma_3\cdot\mathcal{L}_{\textrm{reg}}
\end{split}
\end{align}
We use weights $\gamma_1, \gamma_2, \gamma_3$ and fix the weight on $\mathcal{L}_{\textrm{unobs}}$ to 1.

\paragraph{Baseline models:} In our experiments, we wish to assess the effect of using reports and ratings. Thus, we compare inferences from models with (i) both reports and ratings (\textit{full model}), (ii) only reports (\textit{reports-only model}), and (iii) only ratings (\textit{ratings-only model}). The full model uses both reporting and rating data and all demographic coefficients. Its hyperparameters are set to the specifications listed above. The reports-only is mostly identical to the full model, except it sets a weight of 0 on the loss terms that evaluate against ground-truth ratings $\mathcal{L}_{\textrm{rating}}$. It also learns type specific reporting coefficients $[\theta_k, \alpha_k]$ for all types (not just types with observed ratings) using all datapoints (not just datapoints with observed ratings, since no datapoints have observed ratings!). The ratings-only is mostly identical to the full model, except it sets a weight of 0 on the loss terms that evaluates against ground-truth reports $\mathcal{L}_{\textrm{unobs}}, \mathcal{L}_{\textrm{obs}}$.

\paragraph{Subsampled experiments:} On both semi-synthetic and real data, we run subsampled experiments to compare the ratings-only model to the full URBAN model. We subsample ratings for each type individually. In Figure \ref{fig:semisynthetic_results} (b) and Figure \ref{fig:real_results} (b) we report results for type $k$ after subsampling ratings \textit{only} for type $k$. One of the challenges of sparsity is that as ratings for type $k$ become more sparsely observed, models learn noisier estimates of the true reporting coefficients $\theta_k, \alpha_k$. We address this challenge by (i) using two additional loss components and (ii) modifying how the model learns the reporting coefficients $\alpha_k,\theta_k$.

\begin{table*}[t]
\vspace{-0em}
\centering
\begin{tabular}{l r r}
\hline
\textbf{Type frequency percentile} & \multicolumn{2}{c}{\textbf{Correlation of ratings}} \\ 
& \textbf{Full model} & \textbf{Reports-only model} \\ 
\hline
0-20\%             & $0.269\pm0.025$ & $-0.074\pm0.035$  \\
20-40\%            & $0.197\pm0.021$ & $-0.066\pm0.021$  \\
40-60\%            & $0.367\pm0.011$ & $0.326\pm0.007$   \\
60-80\%            & $0.602\pm0.005$ & $0.581\pm0.004$   \\
80-100\%           & $0.768\pm0.002$ & $0.745\pm0.002$   \\
\hline
\end{tabular}
\caption{\textit{The URBAN model can learn across types.} We evaluate our model's performance in predicting ratings across type frequencies. We measure type frequency as $\bbE_{it}\left[\Tikt\right]$. Across all type frequency quintiles, and particularly for rare types (bottom quintiles), compared to the reports-only model, our full model, which uses both reporting and rating data, predicts more correlated ratings. We show results for all types that the model \textit{does not} observe ratings for, and so the ratings-only model cannot be used at all. We report the mean and 95\% CI across 20 synthetic datasets.}
\vspace{-1.5em}
\label{tab:semisynthetic_ratings_unobserved}
\end{table*}

As a recap, we train all of our models with a loss function made up of four components: (i) report loss for data points with unobserved ratings, (ii) report loss for data points with observed ratings, (iii) rating loss, and (iv) regularization loss. For the models run on subsampled data, we also use:

(v) \textit{Demographic coefficient regularization loss:} $L^2$ norm of the predicted demographic coefficients $\theta_k$ for the type $k$ with subsampled ratings.
\begin{align}
\label{eq:regularization_loss}
\begin{split}
\mathcal{L}_{\theta\textrm{-reg}} &= L^2\left(\theta_k\right)
\end{split}
\end{align} 

As ratings become increasingly sparse, we approach the setting of the reports-only model (where no ratings are observed). The reports-only model estimates ratings by using the estimated probability of observing a report $\hat{P}(T_{ikt})$ as a proxy for the rating. This loss component encourages a model to emulate the reports-only model by forcing the model's predicted ratings $\hat{r}_{ikt}$ to behave similarly to the model's predicted probabilities of a report  $\hat{P}(T_{ikt})$. In particular, this loss constrains the model's ranking of the predicted ratings to match the model's ranking of the predicted probabilities of a report by encouraging each predicted rating to be a scalar mulitple of the predicted probability of a report (and not to additionally depend on demographics). 

(vi) \textit{ReLU penalty on rating coefficient:} Rectified linear unit (ReLU) penalty on the predicted rating coefficient $\alpha_k$ for the type $k$ with subsampled ratings. 
\begin{align}
\label{eq:regularization_loss}
\begin{split}
\mathcal{L}_{\alpha\textrm{-relu}} &= \textrm{ReLU}\left(\alpha_k\right)
\end{split}
\end{align} 

This loss component helps the model learn the inverse relationship (negative $\alpha_k$ coefficient) between ratings and reports -- neighborhoods with lower ratings have worse conditions and therefore receive more reports. When ratings are sparsely observed, this relationship can be difficult to identify without this loss component. 

Thus the overall loss is as follows:
\begin{align}
\label{eq:overall_loss}
\begin{split}
\mathcal{L} &= \mathcal{L}_{\textrm{unobs}} + \gamma_1\cdot\mathcal{L}_{\textrm{obs}}+ \gamma_2\cdot\mathcal{L}_{\textrm{rating}} + \gamma_3\cdot\mathcal{L}_{\textrm{reg}} \\
&+ \gamma_4\cdot\mathcal{L}_{\theta\textrm{-reg}} + \gamma_5\cdot\mathcal{L}_{\alpha\textrm{-relu}}
\end{split}
\end{align}
We use weights $\gamma_1, \gamma_2, \gamma_3, \gamma_4, \gamma_5$ and fix the weight on $\mathcal{L}_{\textrm{unobs}}$ to 1.

For all models run on the full dataset, we use $\gamma_4=\gamma_5=0$ (we use the $\gamma_1,\gamma_2,\gamma_3$ values specified earlier). For the ratings-only models run on subsampled data, we use the same hyperparameters as the ratings-only model run on the full dataset. Thus we use $\gamma_1 = 0, \gamma_2 = 1, \gamma_3 = 10^{-6}, \gamma_4 = 0 \gamma_5 = 0$. For the semi-synthetic experiments, the full URBAN model run on subsampled data uses $\gamma_1=20, \gamma_2=10, \gamma_3=10^{-6}, \gamma_4 = 0 \gamma_5 = 0$. For the real data experiments, the full model run on subsampled data uses $\gamma_1=20, \gamma_2=10, \gamma_3=10^{-6}, \gamma_4 = 0.1 \gamma_5 = 0.1$. Note that we only apply the two additional loss components for the full model run on subsampled real data. All other hyperparameters are identical to the standard model. 

\begin{table*}[]
    \centering
    \renewcommand{\arraystretch}{1.5} 
    \vspace{-0em}
    \begin{tabularx}{\textwidth}{l
                                 >{\centering\arraybackslash}X 
                                 >{\centering\arraybackslash}X 
                                 >{\centering\arraybackslash}X}
    \toprule[1.5pt]
     & \textbf{Full model} & \textbf{Ratings-only model} & \textbf{Reports-only model} \\ 
    \midrule[1.5pt]
    \textbf{Correlation} & $\mathbf{0.5303 \pm 0.0194}$ & $0.5223 \pm 0.0185$ & $0.0993 \pm 0.0143$ \\ 
    \textbf{RMSE} & $\mathbf{0.5833 \pm 0.0123}$ & $0.5852 \pm 0.0116$ & -- \\ 
    \midrule
    \textbf{Top-5 coverage} & $0.1169 \pm 0.0287$ & $\mathbf{0.1262 \pm 0.0232}$ & $0.0123 \pm 0.0137$ \\
    \textbf{Top-10 coverage} & $\mathbf{0.1692 \pm 0.0257}$ & $\mathbf{0.1692 \pm 0.0253}$ & $0.0262 \pm 0.0150$ \\
    \textbf{Top-20 coverage} & $\mathbf{0.2115 \pm 0.0189}$ & $0.2046 \pm 0.0203$ & $0.0354 \pm 0.0069$ \\
    \textbf{Top-50 coverage} & $\mathbf{0.2655 \pm 0.0148}$ & $\mathbf{0.2655 \pm 0.0164}$ & $0.0849 \pm 0.0103$ \\
    \midrule
    \textbf{ECE} & $\mathbf{0.2072 \pm 0.0080}$ & $0.2094 \pm 0.0082$ & -- \\
    \midrule
    \textbf{Correlation gap for low income} & $\mathbf{-0.0028 \pm 0.0030}$ & $-0.0056 \pm 0.0046$ & $-0.0032 \pm 0.0091$ \\
    \textbf{Correlation gap for middle income} & $\mathbf{0.0023 \pm 0.0028}$ & $0.0029 \pm 0.0032$ & $-0.0294 \pm 0.0067$ \\
    \textbf{Correlation gap for high income} & $\mathbf{0.0056 \pm 0.0042}$ & $0.0099 \pm 0.0074$ & $0.0178 \pm 0.0047$ \\
    \textbf{Correlation gap for predominantly minority} & $\mathbf{0.0005 \pm 0.0043}$ & $-0.0028 \pm 0.0055$ & $-0.0214 \pm 0.0135$ \\
    \textbf{Correlation gap for mixed} & $\mathbf{0.0121 \pm 0.0081}$ & $0.0193 \pm 0.0082$ & $-0.0451 \pm 0.0142$ \\
    \textbf{Correlation gap for predominantly white} & $\mathbf{0.0006 \pm 0.0048}$ & $-0.0009 \pm 0.0054$ & $0.0191 \pm 0.0092$ \\
    \midrule
    \textbf{ECE gap for low income} & $\mathbf{-0.0030 \pm 0.0030}$ & $-0.0036 \pm 0.0044$ & $-0.0182 \pm 0.0040$ \\
    \textbf{ECE gap for middle income} & $\mathbf{0.0012 \pm 0.0049}$ & $-0.0049 \pm 0.0046$ & $-0.0109 \pm 0.0036$ \\
    \textbf{ECE gap for high income} & $-0.0061 \pm 0.0040$ & $\mathbf{-0.0022 \pm 0.0048}$ & $-0.0233 \pm 0.0015$ \\
    \textbf{ECE gap for predominantly minority} & $\mathbf{-0.0006 \pm 0.0041}$ & $-0.0040 \pm 0.0059$ & $-0.0092 \pm 0.0035$ \\
    \textbf{ECE gap for mixed} & $\mathbf{-0.0058 \pm 0.0069}$ & $-0.0124 \pm 0.0056$ & $-0.0072 \pm 0.0059$ \\
    \textbf{ECE gap for predominantly white} & $-0.0083 \pm 0.0059$ & $\mathbf{-0.0023 \pm 0.0050}$ & $-0.0332 \pm 0.0037$ \\
    \midrule
    \textbf{Representation ratio (income)} & $0.8945 \pm 0.0070$ & $\mathbf{0.8956 \pm 0.0079}$ & $1.1395 \pm 0.0276$ \\
    \textbf{Representation ratio (race)} & $0.8451 \pm 0.0155$ & $\mathbf{0.8466 \pm 0.0121}$ & $1.1656 \pm 0.0439$ \\
    \bottomrule[1.5pt]
    \end{tabularx}
    \caption{\textit{Real data results.} For the real data, we evaluate rating predictions on additional metrics. We compare our full URBAN model (which uses both reporting and rating data) to a reports-only and a ratings-only model. Our full model outperforms both baseline models on almost all of the metrics (the best performing model is bolded). We report the mean and 95\% CIs over all contiguous two year periods between 2021 and 2023.}
    \vspace{-1.5em}
    \label{tab:real_data_results}
\end{table*}

\begin{figure*}[t]
    \begin{center}
    \includegraphics[width=0.75\textwidth]{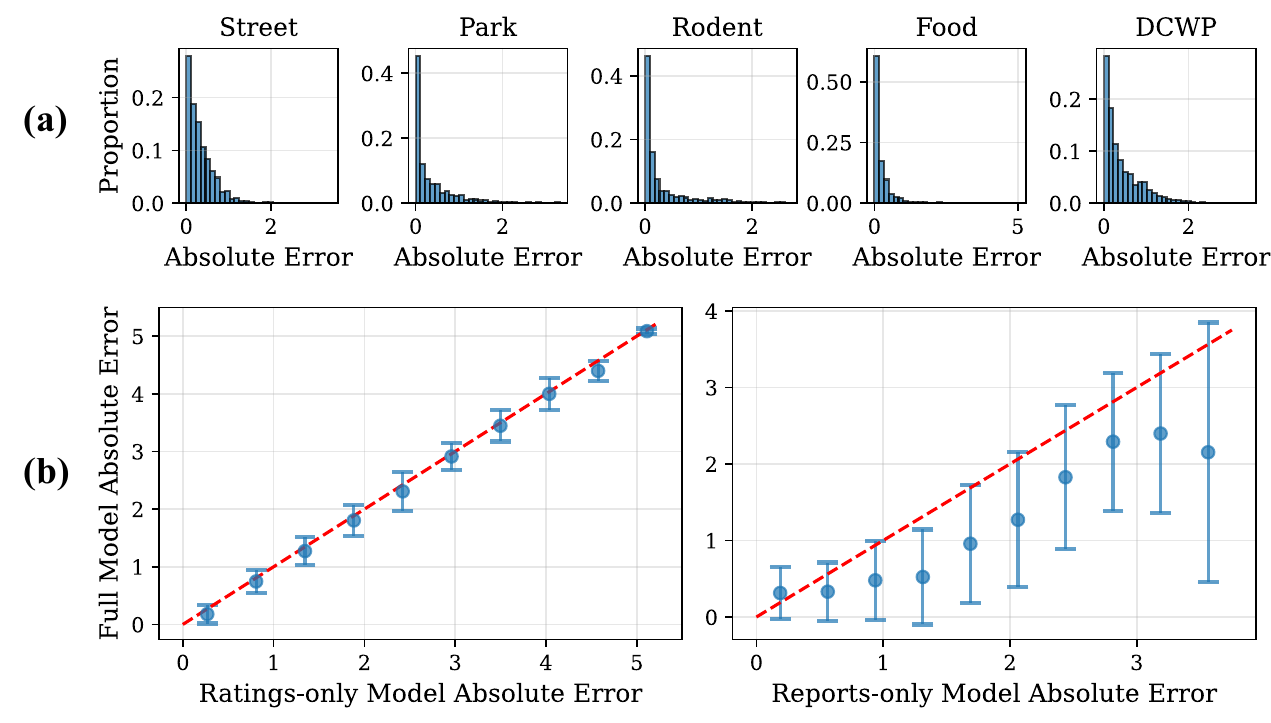}
  \end{center}
    \caption{\textit{Node-level results.} For both our URBAN model and baselines, performance varies significantly across nodes. In (a) we plot the distribution of errors across all nodes. In (b) we compare the error of our full model's predicted ratings across nodes to the error of our baseline methods' predicted ratings. We find that the full model has less error than the reports-only model and comparable error to the ratings-only model. Panel (a) shows results for one model and panel (b) shows results over all contiguous two-year periods between 2021 and 2023.}
    \label{fig:node}
    \vspace{-1.5em}  
\end{figure*}

For the full model run on real subsampled data, we also modify how the model learns the reporting coefficients $\alpha_k,\theta_k$. Our standard full model only learns the reporting coefficients $\alpha_k,\theta_k$ for each type $k$ with observed ratings using datapoints with observed ratings (see Case 1 in \S\ref{sec:model}). However, as ratings become more sparsely observed, we learn the reporting coefficients from fewer datapoints and therefore learn noisier estimates of $\alpha_k,\theta_k$. On the other hand, the reports-only model, which observes no ratings,  learns reporting coefficients $\alpha_k,\theta_k$ for all types $k$ using all datapoints. While this introduces identifiability issues (since we are simulatenously learning ratings $\rikt$ and the coefficient on rating $\alpha_k$), this improves our estimate of the demographic coefficents $\theta_k$ which can now be learned from the entire dataset (not just from datapoints with observed ratings). We use this intuition to modify our full model. We learn reporting coefficients for the type $k$ with subsampled ratings using a weighted combination of gradients from datapoints with observed ratings and datapoints with unobserved ratings. We downweight gradients from datapoints with unobserved ratings using a weight of $0.6$. 

\paragraph{Hyperparameter search:} We conduct a hyperparameter search over the loss weights $\gamma_1, \gamma_2, \gamma_3, \gamma_4, \gamma_5$, embedding dimension sizes, number of layers, batch size, and learning rate using Weights and Biases on a validation set unseen at train or test time. We select the set of hyperparameters that maximize the correlation of predicted reports and ratings. We sweeped over the following hyperparameter values: $\gamma_1\in\{1, 20, 100\}, \gamma_2\in\{1, 20, 100\}, \gamma_3\in\{10^{-1}, 10^{-2}, 10^{-3}, 10^{-4}, 10^{-5}, 10^{-6}, 10^{-7}\}, \gamma_4\in\{0.1, 1, 10, 100, 1000\}, \gamma_5\in\{0.1, 1, 10, 100, 1000\}$, gradient weight on datapoints with unobserved ratings in \{0, 0.2, 0.4, 0.6, 0.8, 1\}, intermediate embedding dimension in \{50, 100, 500, 1000, 2292\}, final embedding dimension in \{50, 100, 500, 1000, 2292\}, number of convolutional layers in \{1, 2, 3\}, batch size in \{4000, 8000, 16000, 32000\}, and learning rate in \{0.001, 0.01, 0.1\}. We ran the sweep using Weights and Biases on a validation set unseen at train or test time. Based on the hyperparameter search, we run experiments with a learning rate of $0.01$ and a batch size of $16000$. Our full URBAN model uses weights $\gamma_1=20, \gamma_2=1, \gamma_3=10^{-6}, \gamma_4=0, \gamma_5=0$. All experiments were implemented using PyTorch Lightning with a random seed set to 0. All experiments were also conducted on a cluster with access to NVIDIA A100 and A6000 GPUs. Our model can comfortably train on one GPU.
\section{Further details on the semi-synthetic experiments}
\label{sec:appendix_semisynthetic_experiments}
\subsection{Semi-synthetic data}
We generate synthetic inspection ratings $\rikt$ using eq. \eqref{eq:semisynthetic_rating}. We separately generate ratings for the train and test split. For example, for the train split, $\mathbb{E}_{t}(\Tikt)$ is defined as the empirical frequency of $\Tikt$ over all weeks in the train time period. We draw $\alpha_k$ and $\theta_k$ from a Gaussian. The mean of the Gaussian is calculated as follows: We take our real rating data, and separately for each type fit a logisitic regression predicting reports from demographics and the ground-truth rating ($\Tikt\sim X_\node, \rikt$). We set the mean $\alpha_k$ and $\theta_k$ (other than the intercept) to be the average over the coefficients predicted across these type-specific logistic regressions. We set the intercept such that the ratings are zero mean. Thus, our generated and real inspection ratings take on both negative and nonnegative values.

\subsection{Semi-synthetic results} 
We evaluate our predicted reports and ratings using both \textit{correlation} and \textit{root mean squared error (RMSE)}. 

\paragraph{Correlation results:} We evaluate reports by calculating the correlation between our model's predicted probability of a report and the average true report across each node/type pair. In other words, we calculate $\textrm{corr}(\hat{P}(\Tikt), \mathbb{E}_t[\Tikt])$. We evaluate ratings by calculating the correlation between our models predicted rating and the average true rating across each node/type pair. In other words, we calculate $\textrm{corr}(\hat{r}_{\node kt}, \mathbb{E}_t[\rikt])$.

In Table \ref{tab:results}, we calculate the average correlation on reports across all node/type pairs and the average correlation on ratings across node/type pairs with observed ratings. The ratings-only model only predicts ratings, so we cannot evaluate its performance on predicting reports. Similarly, the reports-only model only predicts probabilities of reports. Thus in order to estimate the reports-only model's correlation on ratings we use the predicted probability of a report as a proxy for rating and evaluate $\textrm{corr}(\hat{P}(\Tikt), \mathbb{E}_t[\rikt])$.

We compare our full URBAN model's performance to the reports-only model and the ratings-only model. First we compare to the reports-only model. Figure \ref{fig:semisynthetic_results} (a), shows that compared to the reports-only model, our full model's predicted ratings are more correlated with ground truth ratings for each type with observed ratings. Table \ref{tab:semisynthetic_ratings_unobserved} shows that even for types with completely unobserved ratings, compared to the reports-only model, our full model's predicted ratings are more correlated with ground truth ratings. Next we compare to the ratings-only model. In Figure \ref{fig:semisynthetic_results} (b), we show that for types where reports are predictive of ratings and when ratings are sparsely observed, compared to the ratings-only model, our full model's predicted ratings are more correlated with ground truth ratings. For types with completely unobserved ratings, the ratings-only model cannot make predictions. The ratings-only model can only learn ratings for types with observed ratings.

\paragraph{RMSE results:} We evaluate reports by calculating the RMSE between our models predicted probability of a report and the average true report across each node/type pair. In other words, we calculate $\textrm{RMSE}(\hat{P}(\Tikt), \mathbb{E}_t[\Tikt])$. We evaluate ratings by calculating the RMSE between our model's predicted rating and the average true rating across each node/type pair. In other words, we calculate $\textrm{RMSE}(\hat{r}_{\node kt}, \mathbb{E}_t[\rikt])$.

\begin{table}[t]
\vspace{-0em}
\centering
\begin{tabular}{l r}
\hline
\textbf{Covariate}               & \textbf{Mean coefficient} \\ \hline
Bachelors degree population       & $0.234\pm0.030$                           \\
Households occupied by renter                 & $0.174\pm0.028$                             \\
Median age                         & $0.155\pm0.010$                             \\
log(Population density)             & $0.148\pm0.035$                             \\
White population                   & $-0.072\pm0.013$                            \\
log(Median income)                      & $-0.126\pm0.024$                            \\
True inspection rating                        & $-0.193\pm0.009$                             \\
\hline
\end{tabular}
\caption{\textit{Multivariate reporting coefficients.} We report the average predicted multivariate demographic coefficients across types with observed ratings. The estimated coefficients capture known demographic factors: tracts that are more educated, older, or more dense are more likely to report incidents. We also report the coefficient on the true rating. Tracts that have lower ratings are more likely to be reported. We report the mean coefficients and 95\% CIs over all contiguous two year periods between 2021 and 2023.}
\vspace{-1.5em}
\label{tab:real_multivariate_coefficients}
\end{table}

In Table \ref{tab:results}, we calculate the average RMSE on reports across all node/type pairs and the average RMSE on ratings across node/type pairs with observed ratings. We report the mean RMSE across 20 synthetic datasets. We note that the ratings-only model only predicts ratings. Therefore, we cannot evaluate its performance on predicting reports. Similarly, the reports-only model only predicts probabilities of reports. Therefore, we cannot evaluate its performance on predicting ratings. Note that unlike for correlation, when calculating RMSE, we \textit{cannot} use a proxy for rating (e.g., $\hat{P}(\Tikt)$).

\begin{table*}[]
\centering
\vspace{-0em}
\begin{tabular}{l c c c c}
\hline
Cluster & 0 & 1 & 2 & 3 \\ \hline
Race:Non-Hispanic White            & 29\%  & \textbf{55\%}  & 12\%  & 35\% \\
Race:Asian                         & \textbf{18\%}  & 16\%  & 5\% & \textbf{18\%} \\
Race:African-American              & 19\%  & 8\%  & \textbf{35\%}  & 21\% \\
Households occupied by renter & 59\%  & 72\%  & \textbf{87\%}  & 45\% \\
Bachelors degree      & 33\%  & \textbf{71\%}  & 26\%  & 36\% \\
Population                & 3,900  & \textbf{5,600}  & 5,200  & 2,600 \\
Median income            & 71,000  & \textbf{120,000}  & 47,000  & 74,000 \\
Median age                   & 38  & 37  & 35  & \textbf{40} \\
 \hline
\end{tabular}
\caption{\textit{Clustered nodes are demographically distinct.} We find that the clustering correlates with differences in demographics. All differences between clusters are statistically significant ($p < 0.001$, ANOVA test). The largest value in each row is shown in bold.}
\vspace{-1.5em}
\label{tab:node_cluster_data}
\end{table*}

\paragraph{Discussion of reports-only model's results:} We now discuss the reports-only model's results in more detail. As seen in Table \ref{tab:results}, the reports-only model can predict the probability of a report $P(\Tikt)$ well ($r=0.48$). However, compared to the full URBAN model, the reports-only model predicts ratings that are poorly correlated with ground truth ($r=0.62$ for the full model vs. $0.33$ for the reports-only model). In the semi-synthetic data, ratings are produced by eq. \eqref{eq:semisynthetic_rating}. In order to produce well-calibrated ratings, we must recover the true empirical frequencies $\mathbb{E}_t(\Tikt)$ and the true type-specific demographic coefficient $\theta_k$.\footnote{We \textit{do not} need to recover the type-specific rating coefficient $\alpha_k$ to produce \textit{well-calibrated} ratings, we only need to recover $\alpha_k$ if we want to \textit{exactly} recover ratings.} We claim that the reports-only model is unable to predict well-correlated ratings because it cannot recover the true demographic coefficient $\theta_k$. We show this is true by comparing the reports-only model's rating predictions to ratings calculated according to eq. \eqref{eq:semisynthetic_rating} using the \textit{true} value of $\theta_k$ and with $\mathbb{E}_t(\Tikt)$ equal to the reports-only model's estimates of $P(\Tikt)$. In the latter case, our predicted ratings are better correlated with the true synthetic ratings ($r = 0.53$ vs. $0.33$ for the reports-only model). This improvement is solely due to correcting the value of the demographic coefficient $\theta_k$. We note that despite the improvement, the correlation $r = 0.53$ is still below that of the full model ($0.62$). This is because for data points with observed ratings the full model produces better estimates of $P(\Tikt)$ ($r = 0.57$ for the full model vs. $0.50$ for the reports-only model). Overall we have shown that the reports-only model's ratings are poorly correlated with the ground-truth ratings because it \textit{cannot} recover the true demographic coefficients $\theta_k$. Only a model which observes ratings can accurately recover $\theta_k$, therefore even though the reports-only model can accurately predict $\mathbb{E}_t(\Tikt)$ it is fundamentally limited in its rating predictions.
\begin{figure}
\vspace{-0.5em}
  \begin{center}
    \includegraphics[width=0.45\textwidth]{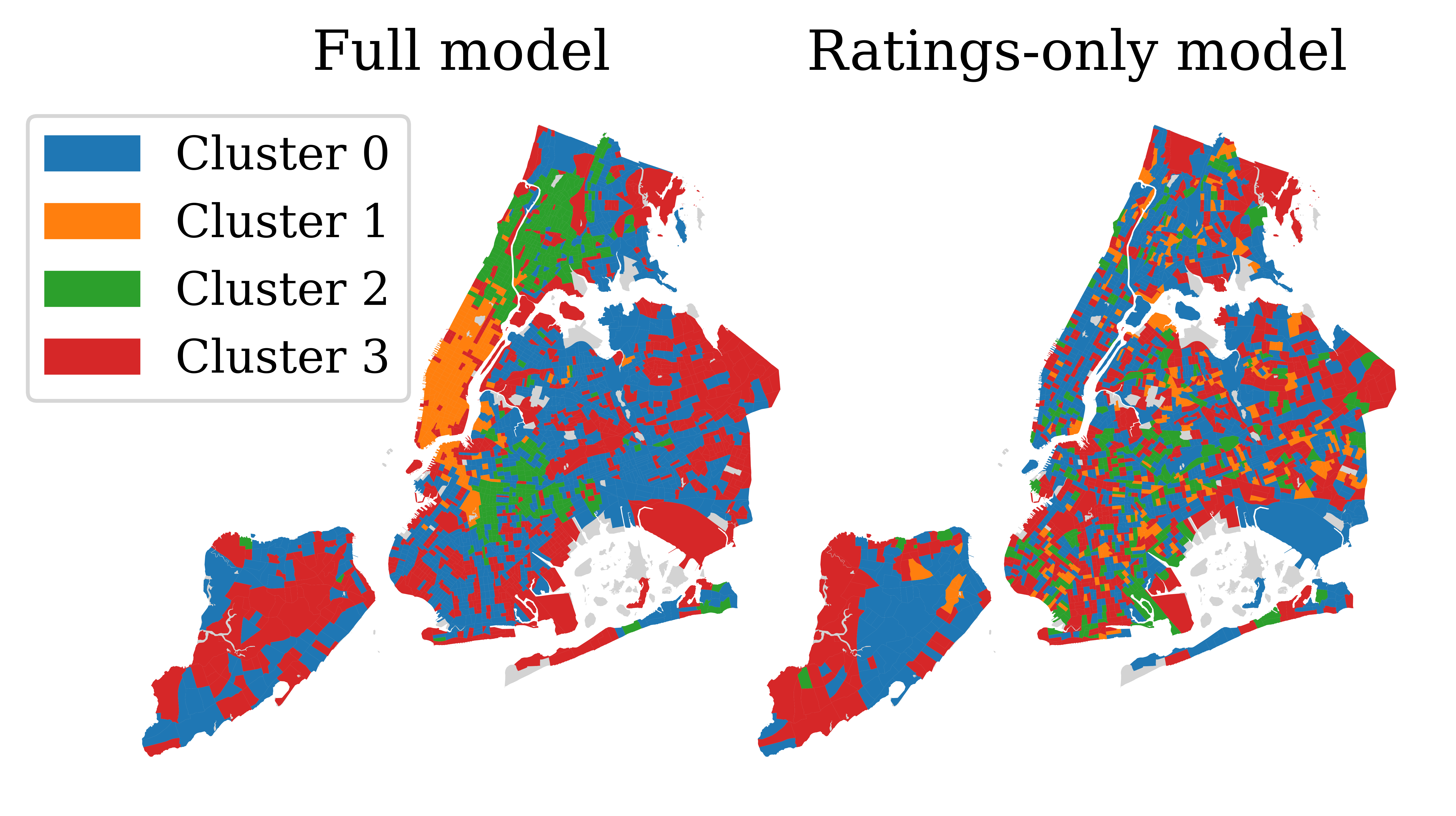}
  \end{center}
  \caption{\textit{Our full model learns spatially correlated ratings.} Using each node's vector of learned ratings over types, we cluster nodes into 5 groups using a $k$ means clustering algorithm. Our model, which learns from both reports and ratings, predicts more spatially clustered ratings than a model which learns only from ratings. }
  \label{fig:node_cluster}
  \vspace{-1.5em}
\end{figure}

\section{Further details on the real-world case study}
\label{sec:appendix_real_data_experiments}

\paragraph{Real data results:} We report our URBAN model's correlation and RMSE on predicted reports and ratings in Table \ref{tab:results}. In Table \ref{tab:real_data_results}, we also evaluate rating predictions using: top-$k$ coverage, expected calibration error (ECE), correlation and ECE gaps for demographic subgroups, and spatial equity under budget constraints as measured by the average demographics of Census tracts which receive resources compared to the average overall demographics. Results are broadly the same across all metrics and our full model generally obtains the best results.

\paragraph{Comparing performance across nodes:} For both our URBAN model and baselines, performance varies significantly across nodes. This is seen in Figure \ref{fig:node} (a) which shows a histogram of performance across nodes. The full model’s mean RMSE for ratings is 0.5833 and the SD across nodes is 0.3922 (comparable magnitude). We find similar results for the ratings-only baseline (mean of 0.5852; slightly higher SD of 0.3959). In Figure \ref{fig:node} (b) we show scatter plots comparing node level performance between our full model and baselines. We find that the full model has less error compared to the reports-only baseline, and has similar error compared to the ratings-only baseline. 

We also analyze which nodes systematically underperform. We find that the top 10\% of Census tracts for which our model's rating predictions have the largest absolute error have a smaller population (3780 vs. population of 3840 for a random set of nodes), lower median income (\$69000 vs. \$72000), lower percentage of the population with a bachelors degree (35\% vs. 36\%), higher percentage of households occupied by a renter (62.7\% vs. 60.9\%), and lower percentage of the population being white (28\% vs 30\%). All of these differences are statistically significant (p-value $<0.05$, two-sample $t$-test). Thus the nodes that underperform represent demographically disadvantaged groups, indicating that the reporting data is in fact demographically biased. The baseline models show similar results. For the ratings-only model, the top 10\% of Census tracts for which the model's rating predictions have the largest absolute error have a smaller population (3800 vs. population of 3840 for a random set of nodes), lower median income (\$69000 vs. \$72000), lower percentage of the population with a bachelors degree (35\% vs. 36\%), higher percentage of households occupied by a renter (62.5\% vs. 60.9\%), and lower percentage of the population being white (28\% vs 30\%). All of the differences except for population are statistically significant (p-value $<0.05$, two-sample $t$-test). For the reports-only model, the top 10\% of Census tracts for which the model's rating predictions have the largest absolute error have a smaller population (3750 vs. population of 3840 for a random set of nodes), lower median income (\$67000 vs. \$72000), lower percentage of the population with a bachelors degree (34\% vs. 36\%), higher percentage of households occupied by a renter (61.4\% vs. 60.9\%), and lower percentage of the population being white (27\% vs 30\%). All of the differences except for percentage of households occupied by a renter are statistically significant (p-value $<0.05$, two-sample $t$-test). Interestingly, the reports-only model worst performing Census tracts have a statistically significantly lower median income, lower percentage of the population with a bachelors degree, and lower percentage of the population being white compared to the full model's worst performing Census tracts (p-value $<0.05$, two-sample $t$-test). This indicates that the full model (which uses both the unbiased rating and biased reporting data) is less affected by demographic bias than the reports-only model (which only uses the biased reporting data).

\begin{figure}[t]
\vspace{-0.5em}
  \begin{center}
    \includegraphics[width=0.28\textwidth]{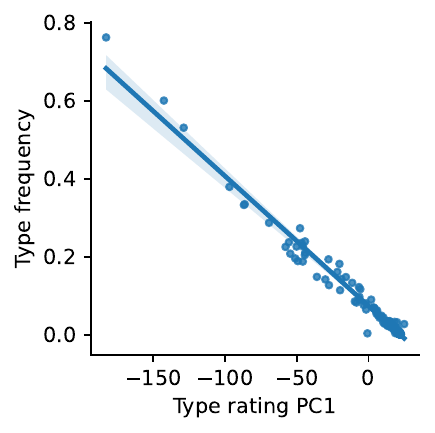}
  \end{center}
  \vspace{-0.5em}
  \caption{\textit{The URBAN model learns interpretable ratings.} Each type's learned ratings over nodes capture type frequency information. In particular, the dimension of highest variance of our type ratings (PC1) has a high correlation with type reporting frequency $\bbE_k\left[\Tikt\right]$.}
  \label{fig:type_rating_frequency}
  \vspace{-1.5em}
\end{figure}

\paragraph{Predicted demographic coefficients:} In Table \ref{tab:real_coefficients} we report the demographic coefficients predicted by univariate models. In Table \ref{tab:real_multivariate_coefficients} we report the demographic coefficients predicted by a multivariate model.

\paragraph{Clustered nodes are demographically distinct:} For each node $\node$, we create a vector $\mathbf{r}_i=\{\rikt\}_{k=1}^{\tau}$ of ratings over all types $k$. We use each node's $\mathbf{r}_i$ vector to cluster the nodes into $4$ groups. We find that the predicted clusters are spatially correlated and demographically distinct. Figure \ref{fig:node_cluster} shows that our clusters are spatially correlated, e.g., there is a clear spatial separation. The clusters correlate with New York City (NYC) borough lines, e.g., Manhattan falls mostly into cluster 1 and the Bronx falls mostly into cluster 2. There are substantial socioeconomic and other demographic differences between Boroughs; the full URBAN model also captures intra-Borough differences, such as that between Midtown Manhattan and Harlem. We also find significant demographic differences between each cluster, and we report the statistically significant differences in Table \ref{tab:node_cluster_data}. Next, we compare our full model clustering to the ratings-only model clustering. Figure \ref{fig:node_cluster} shows that our full model learns more spatially correlated ratings than the ratings-only model.

\paragraph{Clustering ratings for each type:} For each type $k$, we create a vector $\mathbf{r}_k=\{\rikt\}_{i=1}^{n}$ of ratings over all nodes $\node$ to cluster the types into $8$ groups. We find that each group contains a coherent cluster of types, and in Table \ref{tab:type_clusters} we describe and list the types captured by each cluster. Additionally, Figure \ref{fig:type_rating_frequency} shows that the dimension of highest variability (i.e., first PCA dimension) of the $\mathbf{r}_k$ vectors captures type frequency (i.e., $\mathbb{E}_{it}[\Tikt]$).

\begin{table*}[h!]
    \centering
    \caption{\textit{Ratings capture correlations between 311-complaint types.} Using each type's vector of learned ratings over nodes, we cluster types into 8 groups using a $k$ means clustering algorithm. We manually assign a succinct cluster description to each group. We find that the clusters group similar types together.}
    \begin{tabular}{|c p{0.65\textwidth}|} 
\hline
\textbf{Cluster Description} & \textbf{Complaint Types} \\
\hline
\multirow{8}{0.3\textwidth}{\texttt{Housing Maintenance}} 
& \texttt{Appliance (HPD)} \\
& \texttt{Door/Window (HPD)} \\
& \texttt{Electric (HPD)} \\
& \texttt{Flooring/Stairs (HPD)} \\
& \texttt{General (HPD)} \\
& \texttt{Paint/Plaster (HPD)} \\
& \texttt{Plumbing (HPD)} \\
& \texttt{Water Leak (HPD)} 
\\\hline

\multirow{8}{0.3\textwidth}{\texttt{Residential Complaints}} 
& \texttt{Street Light Condition (DOT)} \\
& \texttt{Missed Collection (DSNY)} \\
& \texttt{Heat/Hot Water (HPD)} \\
& \texttt{Unsanitary Condition (HPD)} \\
& \texttt{Blocked Driveway (NYPD)} \\
& \texttt{Illegal Parking (NYPD)} \\
& \texttt{Noise - Residential (NYPD)} \\
& \texttt{Noise - Street/Sidewalk (NYPD)} 
\\\hline

\multirow{9}{0.3\textwidth}{\texttt{Public Space Nuisances}} 
& \texttt{Noise (DEP)} \\
& \texttt{Water System (DEP)} \\
& \texttt{General Construction/Plumbing (DOB)} \\
& \texttt{Derelict Vehicles (DSNY)} \\
& \texttt{Dirty Condition (DSNY)} \\
& \texttt{Illegal Dumping (DSNY)} \\
& \texttt{Abandoned Vehicle (NYPD)} \\
& \texttt{Noise - Commercial (NYPD)} \\
& \texttt{Noise - Vehicle (NYPD)} 
\\\hline

\multirow{12}{0.3\textwidth}{\texttt{Public Health and Landscape}} 
& \texttt{Sewer (DEP)} \\
& \texttt{Building/Use (DOB)} \\
& \texttt{Rodent (DOHMH)} \\
& \texttt{Sidewalk Condition (DOT)} \\
& \texttt{Damaged Tree (DPR)} \\
& \texttt{Dead/Dying Tree (DPR)} \\
& \texttt{New Tree Request (DPR)} \\
& \texttt{Overgrown Tree/Branches (DPR)} \\
& \texttt{Root/Sewer/Sidewalk Condition (DPR)} \\
& \texttt{Dead Animal (DSNY)} \\
& \texttt{Electronics Waste Appointment (DSNY)} \\
& \texttt{Obstruction (DSNY)} 
\\\hline

\multirow{12}{0.3\textwidth}{\texttt{Safety and Mobility Issues}} 
& \texttt{Consumer Complaint (DCA)} \\
& \texttt{Encampment (DHS)} \\
& \texttt{Homeless Person Assistance (DHS)} \\
& \texttt{Elevator (DOB)} \\
& \texttt{Outdoor Dining (DOT)} \\
& \texttt{Traffic Signal Condition (DOT)} \\
& \texttt{Encampment (NYPD)} \\
& \texttt{Non-Emergency Police Matter (NYPD)} \\
& \texttt{Panhandling (NYPD)} \\
& \texttt{For Hire Vehicle Complaint (TLC)} \\
& \texttt{Lost Property (TLC)} \\
& \texttt{Taxi Complaint (TLC)} 
\\\hline
\end{tabular}
    \label{tab:type_clusters}
    \vspace{-3em}
\end{table*}

\begin{table*}[!h]
    \centering
    \vspace{-0em}
    \begin{tabular}{|c p{0.65\textwidth}|} 
\hline
\multirow{30}{0.3\textwidth}{\texttt{Cleanliness and Safety Concerns}}
& \texttt{Asbestos (DEP)} \\
& \texttt{Hazardous Materials (DEP)} \\
& \texttt{Water Quality (DEP)} \\
& \texttt{BEST/Site Safety (DOB)} \\
& \texttt{Electrical (DOB)} \\
& \texttt{School Maintenance (DOE)} \\
& \texttt{Construction Lead Dust (DOHMH)} \\
& \texttt{Face Covering Violation (DOHMH)} \\
& \texttt{Non-Residential Heat (DOHMH)} \\
& \texttt{Standing Water (DOHMH)} \\
& \texttt{Unleashed Dog (DOHMH)} \\
& \texttt{Unsanitary Animal Pvt Property (DOHMH)} \\
& \texttt{Unsanitary Pigeon Condition (DOHMH)} \\
& \texttt{Broken Parking Meter (DOT)} \\
& \texttt{Abandoned Bike (DSNY)} \\
& \texttt{Commercial Disposal Complaint (DSNY)} \\
& \texttt{Dumpster Complaint (DSNY)} \\
& \texttt{Illegal Posting (DSNY)} \\
& \texttt{Litter Basket Complaint (DSNY)} \\
& \texttt{Litter Basket Request (DSNY)} \\
& \texttt{Lot Condition (DSNY)} \\
& \texttt{Snow or Ice (DSNY)} \\
& \texttt{Noise - Helicopter (EDC)} \\
& \texttt{Elevator (HPD)} \\
& \texttt{Bike/Roller/Skate Chronic (NYPD)} \\
& \texttt{Drinking (NYPD)} \\
& \texttt{Drug Activity (NYPD)} \\
& \texttt{Graffiti (NYPD)} \\
& \texttt{Urinating in Public (NYPD)} \\
& \texttt{Taxi Report (TLC)} 
\\\hline

\multirow{27}{0.3\textwidth}{\texttt{Infrastructure and Environmental Health}} 
& \texttt{Consumer Complaint (DCWP)} \\
& \texttt{Air Quality (DEP)} \\
& \texttt{Lead (DEP)} \\
& \texttt{Water Conservation (DEP)} \\
& \texttt{Boilers (DOB)} \\
& \texttt{Emergency Response Team (ERT) (DOB)} \\
& \texttt{Plumbing (DOB)} \\
& \texttt{Real Time Enforcement (DOB)} \\
& \texttt{Special Projects Inspection Team (SPIT) (DOB)} \\
& \texttt{Food Establishment (DOHMH)} \\
& \texttt{Indoor Air Quality (DOHMH)} \\
& \texttt{Smoking (DOHMH)} \\
& \texttt{Curb Condition (DOT)} \\
& \texttt{Street Condition (DOT)} \\
& \texttt{Street Sign - Damaged (DOT)} \\
& \texttt{Street Sign - Dangling (DOT)} \\
& \texttt{Street Sign - Missing (DOT)} \\
& \texttt{Animal in a Park (DPR)} \\
& \texttt{Illegal Tree Damage (DPR)} \\
& \texttt{Maintenance or Facility (DPR)} \\
& \texttt{Violation of Park Rules (DPR)} \\
& \texttt{Graffiti (DSNY)} \\
& \texttt{Residential Disposal Complaint (DSNY)} \\
& \texttt{Sanitation Worker or Vehicle Complaint (DSNY)} \\
& \texttt{Street Sweeping Complaint (DSNY)} \\
& \texttt{Safety (HPD)} \\
& \texttt{Animal-Abuse (NYPD)} 
\\\hline
\end{tabular}
\end{table*}

\begin{table*}[t]
    \centering
    \vspace{-0em}
    \begin{tabular}{|c p{0.65\textwidth}|} 
\hline
\multirow{3}{0.3\textwidth}{} 
& \texttt{Illegal Fireworks (NYPD)} \\
& \texttt{Noise - Park (NYPD)} \\
& \texttt{Traffic (NYPD)} 
\\\hline
\multirow{30}{0.3\textwidth}{\texttt{Facility and Health Inspections}} 
& \texttt{Industrial Waste (DEP)} \\
& \texttt{AHV Inspection Unit (DOB)} \\
& \texttt{Investigations and Discipline (IAD) (DOB)} \\
& \texttt{Scaffold Safety (DOB)} \\
& \texttt{Asbestos (DOHMH)} \\
& \texttt{Beach/Pool/Sauna Complaint (DOHMH)} \\
& \texttt{Day Care (DOHMH)} \\
& \texttt{Drinking Water (DOHMH)} \\
& \texttt{Harboring Bees/Wasps (DOHMH)} \\
& \texttt{Illegal Animal Kept as Pet (DOHMH)} \\
& \texttt{Indoor Sewage (DOHMH)} \\
& \texttt{Mold (DOHMH)} \\
& \texttt{Mosquitoes (DOHMH)} \\
& \texttt{Pet Shop (DOHMH)} \\
& \texttt{Poison Ivy (DOHMH)} \\
& \texttt{Tattooing (DOHMH)} \\
& \texttt{Bike Rack Condition (DOT)} \\
& \texttt{Bus Stop Shelter Complaint (DOT)} \\
& \texttt{Bus Stop Shelter Placement (DOT)} \\
& \texttt{E-Scooter (DOT)} \\
& \texttt{Uprooted Stump (DPR)} \\
& \texttt{Wood Pile Remaining (DPR)} \\
& \texttt{Adopt-A-Basket (DSNY)} \\
& \texttt{Seasonal Collection (DSNY)} \\
& \texttt{Water System (NYC311-PRD)} \\
& \texttt{Disorderly Youth (NYPD)} \\
& \texttt{Noise - House of Worship (NYPD)} \\
& \texttt{For Hire Vehicle Report (TLC)} \\
& \texttt{Green Taxi Complaint (TLC)} 
\\\hline
\end{tabular}
\end{table*}

\end{document}